\documentclass[10pt,twocolumn,letterpaper]{article}

\usepackage{iccv}

\usepackage{times}
\usepackage{epsfig}
\usepackage{graphicx}
\usepackage{amsmath}
\usepackage{amssymb}

\usepackage[caption = false]{subfig}
\usepackage{comment}
\usepackage{color}

\usepackage{dsfont}
\usepackage{multirow}
\usepackage{cite}

\usepackage{array}
\usepackage{booktabs, makecell}

\usepackage{algorithm}
\usepackage{algpseudocode}

\usepackage[pagebackref=true,breaklinks=true,letterpaper=true,colorlinks,bookmarks=false]{hyperref}

\newcolumntype{L}[1]{>{\raggedright\arraybackslash}m{#1}}
\newcolumntype{C}[1]{>{\centering\arraybackslash}m{#1}}
\newcolumntype{R}[1]{>{\raggedleft\arraybackslash}m{#1}}
\newcolumntype{+}{>{\global\let\currentrowstyle\relax}}
\newcolumntype{^}{>{\currentrowstyle}}

\newcommand{\RomNum}[1]{\MakeUppercase{\romannumeral #1}}

\newcommand{\bfr}{\ensuremath{{\mathbf{r}}}}

\newcommand{\bfx}{\ensuremath{{\mathbf{x}}}}

\newcommand{\bfu}{\ensuremath{{\mathbf{u}}}}
\newcommand{\bfv}{\ensuremath{{\mathbf{v}}}}
\newcommand{\bfU}{\ensuremath{{\mathbf{U}}}}

\newcommand{\bfV}{\ensuremath{{\mathbf{V}}}}

\newcommand\notsotiny{\fontsize{6pt}{10pt}\selectfont}

\usepackage[capitalize]{cleveref}
\crefname{section}{Sec.}{Secs.}
\Crefname{section}{Section}{Sections}
\Crefname{table}{Table}{Tables}
\crefname{table}{Tab.}{Tabs.}

\iccvfinalcopy 

\def\iccvPaperID{****} 

\ificcvfinal\fi

\begin{document}

\title{Recursive Video Lane Detection}

\author{Dongkwon Jin\\
Korea University\\
{\tt\small dongkwonjin@mcl.korea.ac.kr}
\and
Dahyun Kim\\
Korea University\\
{\tt\small dhkim@mcl.korea.ac.kr}
\and
Chang-Su Kim\\
Korea University\\
{\tt\small changsukim@korea.ac.kr}
}

\maketitle
\ificcvfinal\thispagestyle{empty}\fi


\begin{abstract}
    A novel algorithm to detect road lanes in videos, called recursive video lane detector (RVLD), is proposed in this paper, which propagates the state of a current frame recursively to the next frame. RVLD consists of an intra-frame lane detector (ILD) and a predictive lane detector (PLD). First, we design ILD to localize lanes in a still frame. Second, we develop PLD to exploit the information of the previous frame for lane detection in a current frame. To this end, we estimate a motion field and warp the previous output to the current frame. Using the warped information, we refine the feature map of the current frame to detect lanes more reliably. Experimental results show that RVLD outperforms existing detectors on video lane datasets. Our codes are available at \href{https://github.com/dongkwonjin/RVLD}{https://github.com/dongkwonjin/RVLD}.
\end{abstract}

\section{Introduction}
\begin{figure}[t]
  \centering
  \includegraphics[width=1\linewidth]{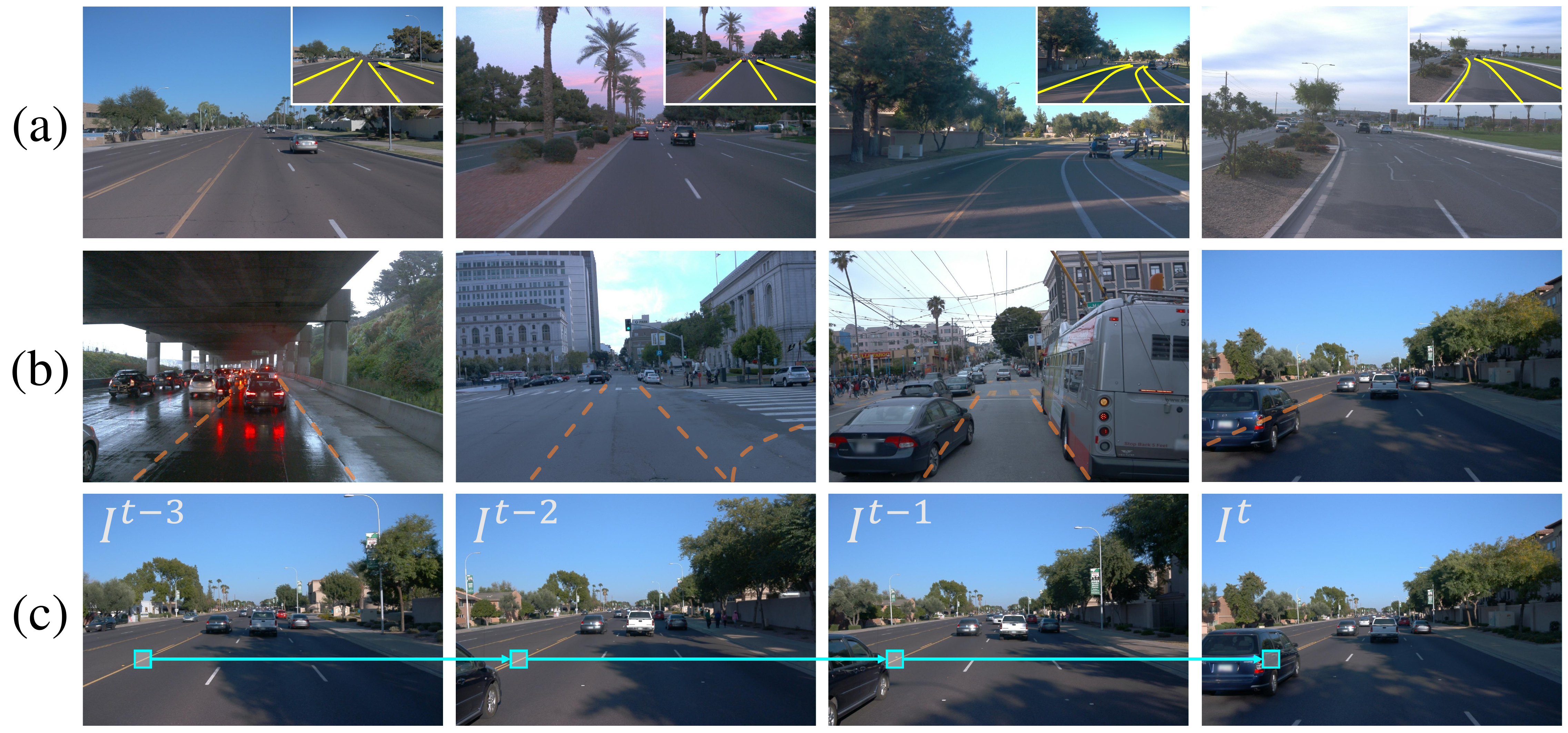}
  \caption{(a) For lane detection, it is required to identify visible lane pixels, represented by yellow lines in the insets. (b) Due to occlusions by nearby vehicles or glistening conditions on wet roads, lanes may be unobvious, which are depicted by orange dotted lines. Besides, lanes may not be marked at crossroads. (c) In a current frame $I^t$, which is also the rightmost one in (b), some lane parts are occluded by a vehicle but visible in past frames. By utilizing visual cues along the temporal axis, depicted by cyan arrows, we can localize the implied lane parts more reliably.}
  \label{fig:opening}
\end{figure}

Lane detection is a task to localize lanes in a road scene, which is crucial for either autonomous or human driving. For lane detection, it is necessary to exploit visual cues of lanes, as illustrated in Figure~\ref{fig:opening}(a). Early methods extract low-level features such as image gradients or colors \cite{he2004,aly2008,hillel2014,zhou2010}. Recently, deep learning techniques have been developed to cope with challenging scenes. Some of them are based on semantic segmentation  \cite{pan2018,zheng2021,qiu2022mfialane,hou2019road,hou2020inter}, in which each pixel is classified into either lane category or not. To ensure lane continuity, several techniques have been proposed, including parametric curve modeling \cite{neven2018,wang2020poly,tabelini2021ICPR,liu2021end,feng2022} and keypoint association \cite{qu2021,wang2022,xu2022}. However, despite yielding promising results, they may fail to detect less visible lanes, as in Figure~\ref{fig:opening}(b). To detect such lanes reliably, anchor-based lane detectors \cite{li2019line,tabelini2021CVPR,jin2022,zheng2022} have been proposed. They generate a set of lane anchors, detect lanes through the binary classification of each anchor, and then regress the detected ones. However, all these methods detect lanes from still images without considering high inter-frame correlation in a video.

In autonomous driving systems, video frames are captured consecutively. Video lane detection aims to detect lanes in such a video by exploiting inter-frame correlation, rather than processing each frame independently. It can detect implied lanes in a current frame more reliably using past information, as shown in Figure~\ref{fig:opening}(c). But, relatively few techniques have been proposed for video lane detection. Zou \etal \cite{zou2019robust} developed a memory network to aggregate the features of past and current frames. Similarly, in \cite{zhang2021lane,zhang2021,tabelini2022,wang2022video}, they combined the features of a current frame with those of past frames and then detected lanes from the mixed features. However, these techniques require several past frames as input and do not reuse the mixed features in subsequent frames.

Recently, the first video lane dataset called VIL-100 \cite{zhang2021} was constructed, containing 100 videos. However, the number of images is only 10K, and most images are collected from highway scenes. Also, OpenLane \cite{chen2022}, a huge dataset for 3D lane detection, was proposed. It consists of 200K images from 1,000 videos and annotates lanes with both 2D and 3D coordinates. But, it is unsuitable for video lane detection because it provides annotations for visible lane parts only: First, the same lane is sometimes broken into multiple parts. Second, some annotations are temporally incoherent because of invisible parts in certain frames. To overcome these issues, we modify OpenLane by filling in missing lane parts semi-automatically based on matrix completion \cite{candes2010power}. The modified dataset is called OpenLane-V.

In this paper, we propose a novel video lane detector called RVLD, which records the state of a current frame and passes it recursively to the next frame to improve detection results. Figure~\ref{fig:Overview} is an overview of the proposed RVLD. First, we design the intra-frame lane detector (ILD) that performs encoding, decoding, and then non-maximum suppression (NMS) to localize lanes in a still image. Second, we develop the predictive --- or inter-frame --- lane detector (PLD) to detect lanes in a current frame using the information in the previous one. Specifically, we estimate a motion field between adjacent frames and warp the previous output to the current frame. Using the warped information, we refine the feature map of the current frame to detect lanes more reliably. Experimental results show that the proposed RVLD outperforms existing techniques on both VIL-100 and OpenLane-V datasets.

This work has the following major contributions:
\begin{itemize}
\itemsep0mm
\item The proposed RVLD improves the detection results in a current frame using a single previous frame only and yields outstanding performances on video datasets.
\item We develop simple yet effective modules for motion estimation and feature refinement to exploit the previous information reliably.
\item We modify the OpenLane dataset to make it more suitable for video lane detection. It is called OpenLane-V.\footnote{OpenLane-V is available at \href{https://github.com/dongkwonjin/RVLD}{https://github.com/dongkwonjin/RVLD}.}
\item We introduce two metrics, flickering and missing rates, for video lane detection.
\end{itemize}

\section{Related Work}

\subsection{Image-Based Lane Detection}
Various techniques have been developed to detect lanes in a still image. Some are based on semantic segmentation \cite{pan2018,zheng2021,qiu2022mfialane,hou2019road,hou2020inter}, in which pixelwise classification is conducted to decide whether each pixel belongs to a lane or not. Pan \etal \cite{pan2018} propagated spatial information through message passing. Zheng \etal \cite{zheng2021} adopted recurrent feature aggregation, while Qiu \etal \cite{qiu2022mfialane} did multi-scale aggregation. Hou \etal \cite{hou2019road} performed self-attention distillation. Moreover, Hou \etal \cite{hou2020inter} employed teacher and student networks. In \cite{qin2020}, Qin \etal determined the location of each lane on selected rows only. Liu \etal \cite{liu2021condlanenet} developed a conditional lane detection strategy based on the row-wise formulation.

In semantic segmentation, the continuity of a detected lane may not be preserved. To maintain the continuity, parametric curve modeling \cite{neven2018,wang2020poly,tabelini2021ICPR,liu2021end,feng2022} and keypoint association \cite{qu2021,wang2022,xu2022} have been developed.
Neven \etal \cite{neven2018} did the polynomial fitting of segmented lane pixels. In \cite{wang2020poly,tabelini2021ICPR}, neural networks were designed to regress polynomial coefficients of lanes. Also, Liu \etal \cite{liu2021end} developed a transformer network to predict cubic lane curves. Feng \etal \cite{feng2022} employed Bezier curves.
In \cite{qu2021}, Qu \etal estimated multiple keypoints and linked them to reconstruct lanes. Wang \etal \cite{wang2022} regressed the offsets from a starting point to keypoints and grouped them into a lane instance. Xu \etal \cite{xu2022} predicted four offsets from each lane point to the two adjacent ones and the topmost and bottommost ones.

\begin{figure}[t]
  \centering
  \includegraphics[width=1\linewidth]{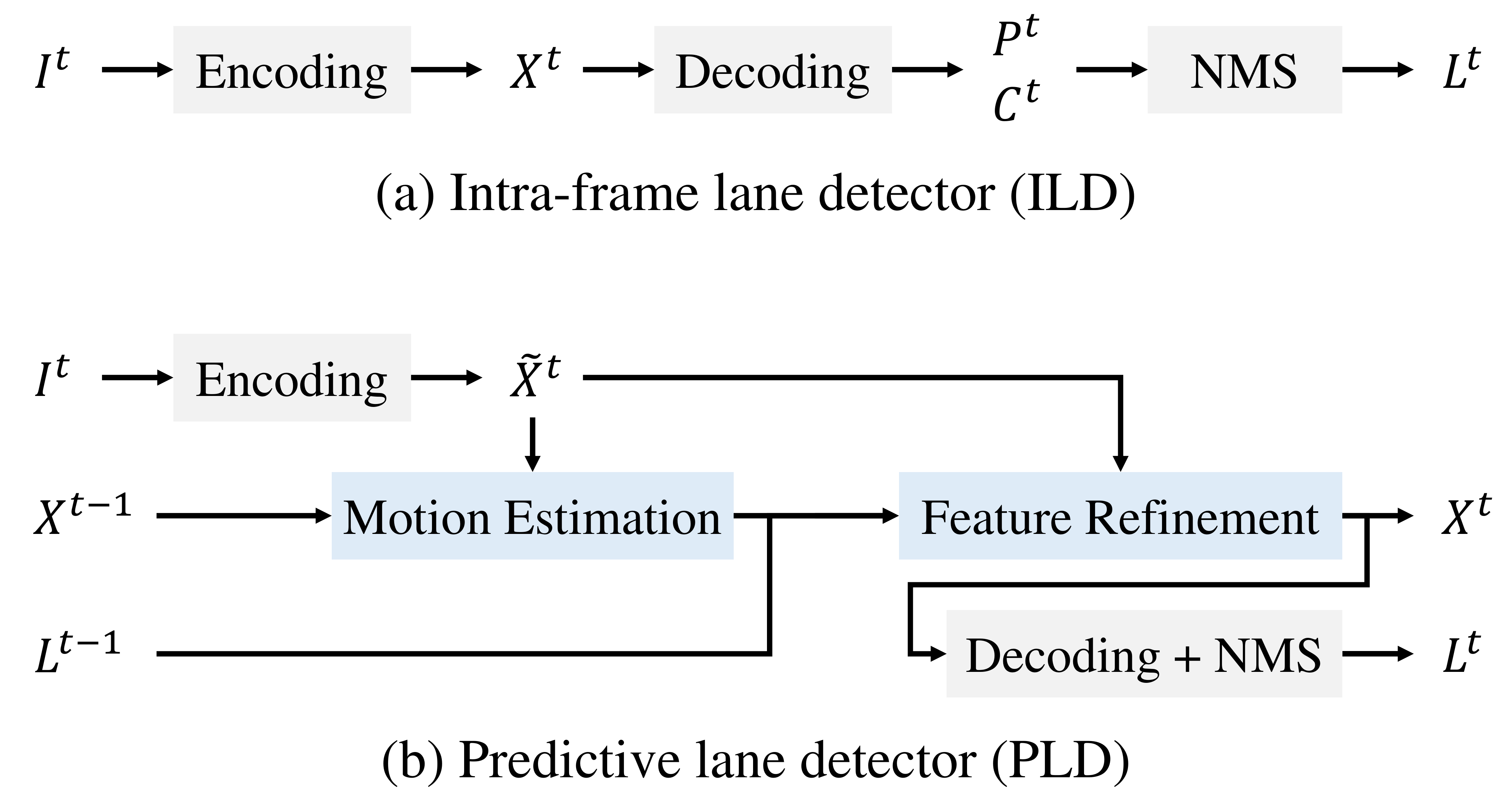}
  \caption{(a) In ILD, we encode a frame $I^t$ at time $t$ into a feature map $X^t$ and decode $X^t$ into a probability map $P^t$ and a coefficient map $C^t$. Then, via NMS, we generate a lane mask $L^t$. (b) In PLD, we utilize previous output $X^{t-1}$ and $L^{t-1}$ to detect lanes in $I^t$. First, we estimate a motion field from $I^t$ to $I^{t-1}$. Then, we backward warp the previous output and refine the feature map $\tilde{X}^t$ of $I^t$ into $X^t$ using the warped information. Lastly, we obtain a lane mask~$L^t$ through the decoding and NMS processes.}
  \label{fig:Overview}
\end{figure}

Meanwhile, the anchor-based detection framework also has been developed \cite{chen2019,li2019line,xu2020curvelane,tabelini2021CVPR,jin2022,zheng2022}.
These techniques generate lane anchors and then classify and regress each anchor by estimating the lane probability and the offset vector. Vertical line anchors were used in \cite{chen2019,xu2020curvelane}. Global features of straight line anchors were extracted to detect lanes in \cite{li2019line,tabelini2021CVPR}. Zheng \etal \cite{zheng2022} extracted multi-scale feature maps and refined them by aggregating global features of learnable line anchors. Also, Jin \etal \cite{jin2022} introduced data-driven descriptors called eigenlanes. They generated lane anchors via a low-rank approximation of a lane matrix to represent lanes compactly and precisely.


\subsection{Video-Based Lane Detection}
There are relatively few video-based lane detectors.
Zou \etal \cite{zou2019robust} and Zhang \etal \cite{zhang2021lane} employed recurrent neural networks to exploit temporal correlation by fusing the features of a current frame with those of several past frames. Zhang \etal \cite{zhang2021} developed a video lane detector using the first video lane dataset VIL-100. They aggregated features of a current frame and multiple past frames based on the attention mechanism \cite{oh2019,vaswani2017} to detect lanes in the current frame. Tabelini \etal \cite{tabelini2022} extracted lane features in video frames, based on the anchor-based detector in \cite{tabelini2021CVPR}, and combined those features. Wang \etal \cite{wang2022video} exploited spatial and temporal information in neighboring video frames by extending the feature aggregation module in \cite{zheng2021}. They also designed loss functions to maintain the geometric consistency of lanes. These techniques require several past frames for feature aggregation but do not reuse the aggregated features in future frames. On the contrary, the proposed algorithm maintains the memory for a single previous frame only but propagates the temporal information recursively to extract lanes in each frame effectively.

\section{Proposed Algorithm}
Given a video sequence, we conduct lane detection in the first frame using ILD in Figure~\ref{fig:Overview}(a) and then do so in the remaining frames using PLD in Figure~\ref{fig:Overview}(b) recursively.

\begin{figure}[t]
  \centering
  \includegraphics[width=1\linewidth]{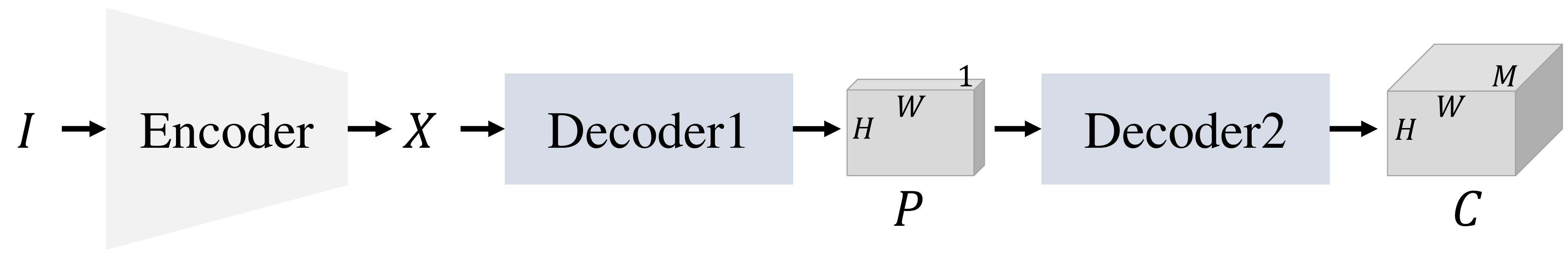}
  \caption{Encoding and decoding in ILD.}
  \label{fig:ILD}
\end{figure}

\begin{figure}[t]
  \centering
  \includegraphics[width=1\linewidth]{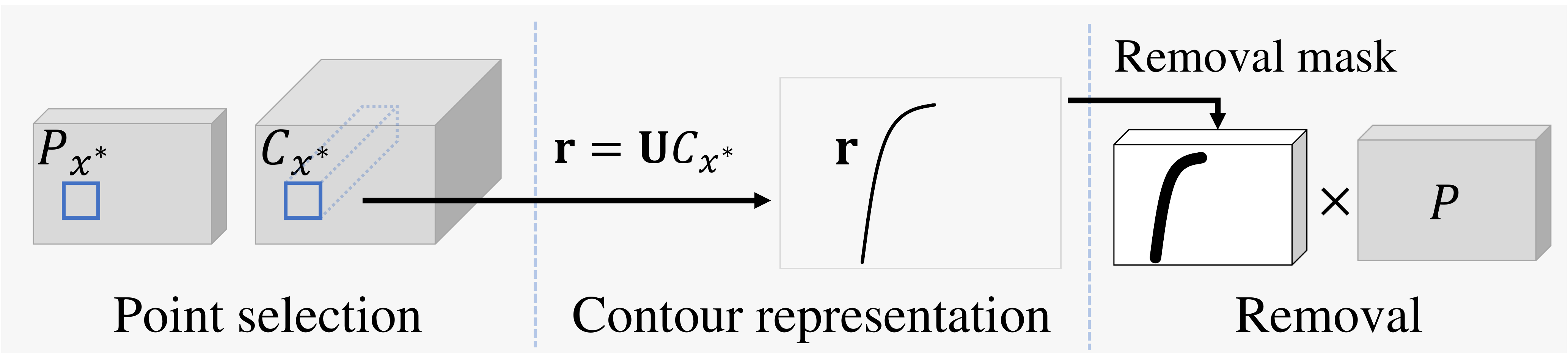}
  \caption{Illustration of NMS in ILD.}
  \label{fig:NMS}
\end{figure}

\subsection{ILD}
\label{ssec:intra}
\noindent\textbf{Encoding and decoding:}
ILD includes a single encoder and two decoders in series, as shown in Figure~\ref{fig:ILD}. Given an image $I$, the encoder extracts a feature map $X \in \mathbb{R}^{H\times W\times K}$ by aggregating multi-scale convolutional features of ResNet18 \cite{he2016deep}. Then, the two decoders sequentially generate a probability map $P \in \mathbb{R}^{H\times W\times 1}$ and a coefficient map $C \in \mathbb{R}^{H\times W\times M}$ by
\begin{equation}\label{eq:intra}
    \textstyle
    P = \sigma(f_1(X)) \,\, \mbox{ and } \,\, C = f_2(P).
\end{equation}
Here, $\sigma(\cdot)$ is the sigmoid function, and $f_1$ and $f_2$ are 2D convolutional layers. Let $x$ denote a pixel coordinate, and $P_x$ be the value of $P$ at $x$. Note that $P_x$ is the probability that pixel $x$ belongs to a lane.

Also, each element in $C$ is a coefficient vector in the $M$-dimensional eigenlane space \cite{jin2022}, in which lanes are represented compactly with $M$ basis vectors. Specifically, $C_x$ is the coefficient vector for a lane containing $x$, if the lane exists. To regress such coefficient vectors more accurately, we use a positional bias map \cite{vaswani2017} as additional input to decoder 2. The architecture of ILD is described in detail in the supplemental document (Section A).

\begin{figure}[t]
  \centering
  \includegraphics[width=1\linewidth]{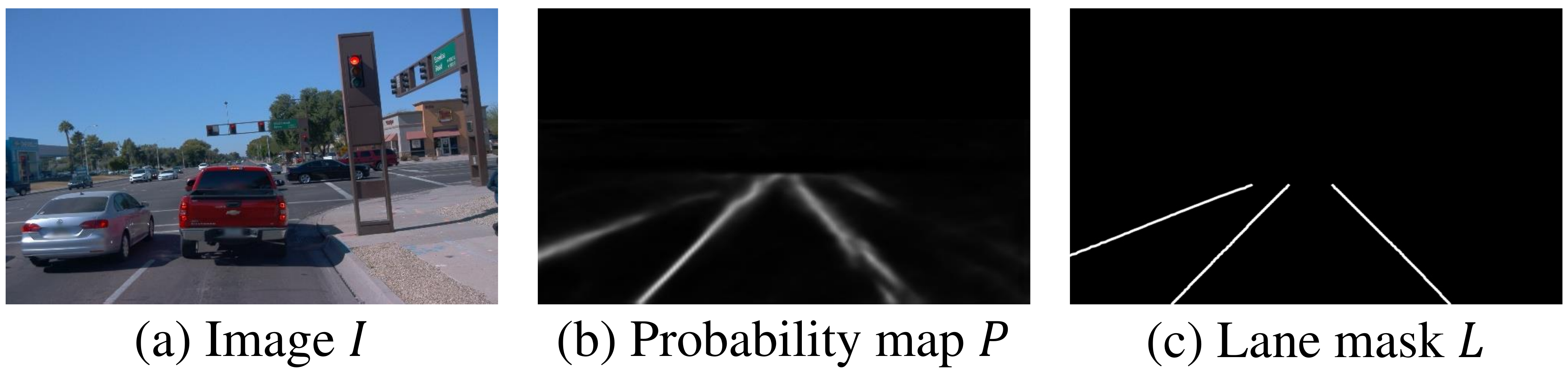}
  \caption{Some lanes are partly occluded in (a). While the probability map $P$ does not represent the invisible parts reliably in (b), the lane mask $L$ reconstructs continuous lanes clearly in (c).}
  \label{fig:ILD_results}
  \vspace*{-0.2cm}
\end{figure}

\vspace*{0.1cm}
\noindent\textbf{NMS:}
Using the probability map $P$ and the coefficient map $C$, we determine reliable lanes through NMS, as illustrated in Figure~\ref{fig:NMS}. First, we select the optimal pixel $x^*$ with the highest probability in $P$. Then, we reconstruct the corresponding lane $\bfr$ by linearly combining $M$ eigenlanes with the coefficient vector $C_{x^*}$, which is given by
\begin{equation} \label{eq:x_approx}
\bfr = \bfU C_{x^*} = [\bfu_1, \cdots, \bfu_M] C_{x^*}
\end{equation}
where $\bfu_1, \cdots, \bfu_M$ are the $M$ eigenlanes \cite{jin2022}. Note that $\bfr$ is a column vector composed of the horizontal coordinates of lane points, which are sampled vertically. By dilating the lane curve $\bfr$, we construct a removal mask, as in Figure~\ref{fig:NMS}, and prevent the pixels within the mask from being selected in the remaining iterations. We iterate this NMS process until $P_{x^*}$ is higher than 0.5.

Finally, using the selected lanes, ILD outputs a binary lane mask $L \in \mathbb{R}^{H\times W\times 1}$: $L_x = 1$ if $x$ belongs to a lane, and $L_x=0$ otherwise. Figure~\ref{fig:ILD_results} illustrates that the lane mask $L$ reduces the ambiguity in the probability map $P$ and reconstructs continuous lanes effectively.

\subsection{PLD}
\label{ssec:inter}

\noindent\textbf{Problem formulation:}
We aim to detect lanes in a current frame $I^t$ using the information in the past frame $I^{t-1}$, as in Figure~\ref{fig:Overview}(b). Note that $X^{t-1}$ and $L^{t-1}$ are the feature map and the lane mask of $I^{t-1}$, respectively. Let $\tilde{X}^t$ be the feature map of $I^t$ obtained by the encoder. Both ILD and PLD use the same encoder, but in PLD we refine $\tilde{X}^t$ to $X^t$ by exploiting $X^{t-1}$ and $L^{t-1}$. Then, from the refined feature $X^t$, we obtain a more reliable lane mask $L^t$. The results $X^t$ and $L^t$ are, in turn, used to detect lanes in the future frame $I^{t+1}$. Notice that PLD is a first-order Markov chain \cite{winston2004}, in which the future outcome at time $t+1$ is influenced by the current state at time $t$ only and independent of the past states at times less than or equal to $t-1$. To perform this recursive detection reliably, we develop simple yet effective modules for motion estimation and feature refinement. The detailed structure of PLD is in the supplement (Section A).

\begin{figure}[t]
  \centering

  \includegraphics[width=1\linewidth]{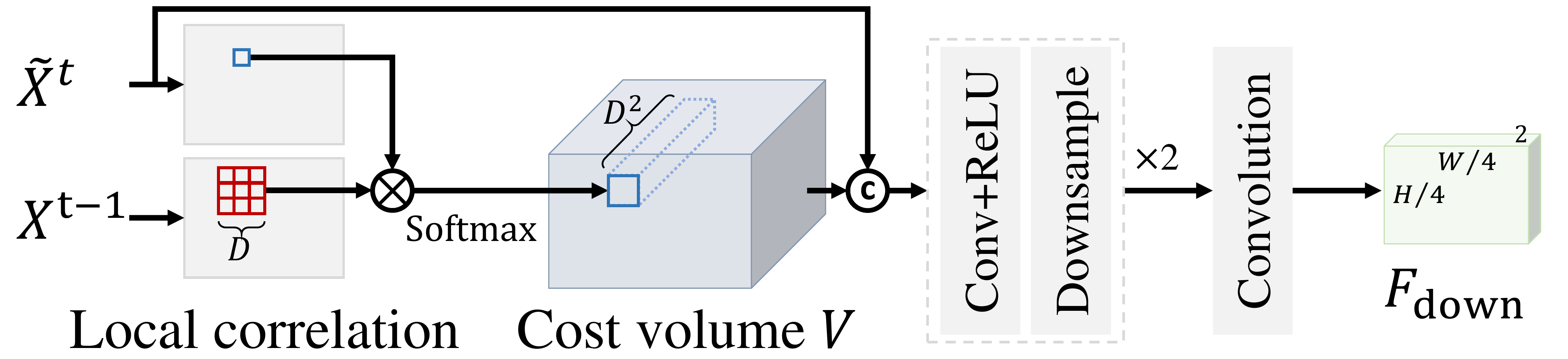}

  \caption{The structure of the proposed motion estimator: Given feature maps $\tilde{X}^t$ and $X^{t-1}$ of adjacent frames $I^t$ and $I^{t-1}$, we estimate a down-sampled motion field $F_\textrm{down}$ from $I^t$ to $I^{t-1}$.}
  \label{fig:motion_estimator}
\end{figure}

\vspace*{0.1cm}
\noindent\textbf{Motion estimation:}
We estimate a motion field from the current frame $I^t$ to the past frame $I^{t-1}$ by designing a lightweight motion estimator in Figure~\ref{fig:motion_estimator}. The motion estimator takes feature maps $\tilde{X}^t$ and $X^{t-1}$ as input. Then, it estimates a down-sampled motion field $F_\textrm{down} \in \mathbb{R}^{H/4 \times W/4\times 2}$, where each element indicates a 2D motion vector from $I^t$ to $I^{t-1}$. To this end, as in \cite{sun2018pwc}, we construct a cost volume $V \in \mathbb{R}^{H \times W\times D^2}$ by first computing the local correlation
\begin{equation}
    \textstyle
    {\bar V}_x(d) =  ({X^{t-1}_{x+d}})^\top \tilde{X}^t_x
\end{equation}
where $d$ is a displacement vector within the search window ${\cal D}=[-s, s] \times [-s, s]$. Thus, we compute $|{\cal D}|=D^2$ correlations for each $x$, where $D=2s+1$. Then, we obtain the cost volume $V$ through $V_x = {\rm softmax}({\bar V}_x)$. Each element in $V_x$ informs of the similarity between $x$ and a matching candidate. We concatenate the cost volume $V$ with $\tilde{X}^t$ and then apply convolutions and down-sampling operations to obtain the motion field $F_\textrm{down}$.

\begin{figure}[t]
  \centering
  \includegraphics[width=1\linewidth]{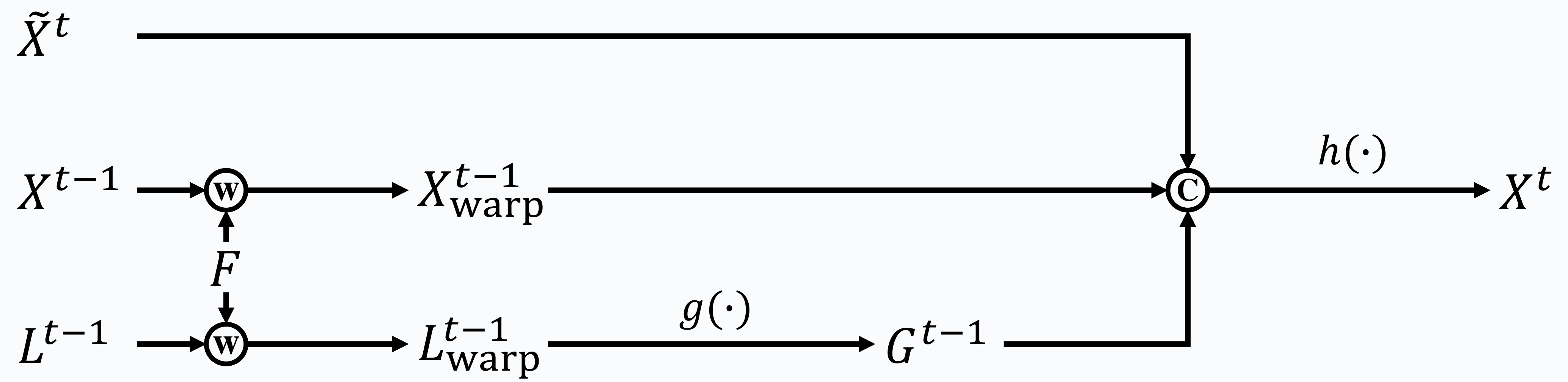}
  \caption{A diagram of the feature refinement module.}
  \label{fig:refinement}
\end{figure}

\vspace*{0.1cm}
\noindent\textbf{Feature refinement:}
Some lanes may be implied or even invisible in $I^t$ due to occlusions, lack of light, or weather conditions. It is difficult to detect such lanes using the contextual information in the current frame only. In other words, $\tilde{X}^t$ of $I^t$ is not sufficiently informative. Thus, we refine $\tilde{X}^t$ into $X^t$ by exploiting the past information in $I^{t-1}$.

In Figure~\ref{fig:refinement}, we first warp the previous output $X^{t-1}$ and $L^{t-1}$ to the current frame by
\begin{equation}
\textstyle
\begin{gathered}
    X^{t-1}_\textrm{warp} = \phi_B(X^{t-1}, F), \;\;\; L^{t-1}_\textrm{warp} = \phi_B(L^{t-1}, F),
\end{gathered}
\end{equation}
where $F \in \mathbb{R}^{H \times W\times 2}$ is the up-sampled motion field of $F_\textrm{down}$ via bilinear interpolation, and $\phi_{B}$ is the backward warping operator \cite{wolberg1990invwarping}. Then, from the warped lane mask $L^{t-1}_\textrm{warp}$, we obtain a guidance feature map by
\begin{equation}
    \textstyle
    G^{t-1} = g(L^{t-1}_\textrm{warp})
\end{equation}
where $g$ is composed of 2D convolutional layers to increase the channel dimension to $K$. Notice that $L^{t-1}_\textrm{warp}$ preserves the structural continuity of lanes similarly to Figure~\ref{fig:ILD_results}(c). $G^{t-1}$ also contains such information. Thus, we may deliver the continuous lane information even for partially occluded lanes to $I^t$. Consequently, we produce the refined feature map $X^t$ by aggregating $G^{t-1}$, $X^{t-1}_\textrm{warp}$, and $\tilde{X}^t$,
\begin{equation}
\textstyle
    X^t = h([G^{t-1}, X^{t-1}_\textrm{warp}, \tilde{X}^t])
    \label{eq:agg_refine}
\end{equation}
where $[\cdot]$ denotes the channel-wise concatenation, and $h$ consists of 2D convolutional layers to reduce the channel dimension back to $K$.

Then, using the refined feature $X^t$, we detect a more reliable lane mask $L^t$ by performing the decoding and NMS processes in the same way as ILD, as shown in Figure~\ref{fig:Overview}.

\subsection{Training}

We define the loss for training ILD as
\begin{equation}\label{eq:ILDloss}
    \textstyle
    \ell_{\rm ILD} = \ell_{\rm cls}(P_x, \bar{P}_x)+\ell_{\rm reg}(C_x, \bar{C}_x)
\end{equation}
where $P_x$ and $C_x$ are the output of ILD in \eqref{eq:intra}, and $\bar{P}_x$ and $\bar{C}_x$ are their ground-truth (GT), respectively. Also, $\ell_{\rm cls}$ is the focal loss \cite{lin2017focal} over binary classes, and $\ell_{\rm reg}$ is the LIoU loss \cite{zheng2022} between a predicted lane contour $\bfr$ in \eqref{eq:x_approx} and its ground-truth $\bar{\bfr}$.

For PLD, in addition to the loss in (\ref{eq:ILDloss}), we employ a loss function
\begin{equation}
\ell_{\rm flow} = \| \bar{P}^{t-1}_\textrm{warp}-\bar{P}^{t} \|^2 = \| \phi_B (\bar{P}^{t-1}, F) -\bar{P}^{t} \|^2
\end{equation}
to train the motion estimation module. In other words, it is trained to yield a motion field $F$ such that the warped result of the ground-truth probability map $\bar{P}^{t-1}$ is close to $\bar{P}^{t}$.

We first train ILD and then, after fixing it, train the motion estimation and feature refinement modules in PLD from scratch. The results of PLD in a current frame are, in turn, used in the next frame. Hence, for stable training of PLD, we compose three consecutive frames as a training unit. The training process and hyper-parameters are detailed in the supplement (Section A).

\captionsetup[subfigure]{labelformat=empty}
\begin{figure}[t]
\vspace{-0.3cm}

    \begin{flushright}
    \subfloat {\raisebox{1.3em}{\rotatebox[origin=t]{90}{\scriptsize OpenLane}}}\hspace{-0.01cm}\,\!
    \subfloat {\includegraphics[width=1.98cm,height=1.3cm]{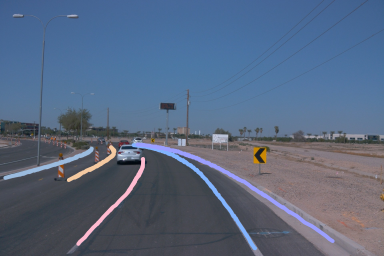}}\,\!\!
    \subfloat {\includegraphics[width=1.98cm,height=1.3cm]{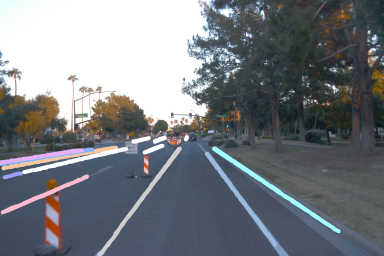}}\,\!\!
    \subfloat {\includegraphics[width=1.98cm,height=1.3cm]{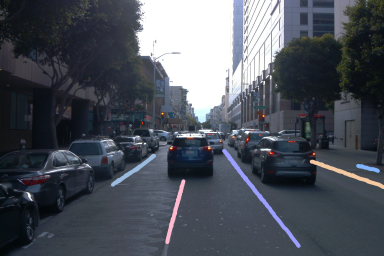}}\,\!\!
    \subfloat {\includegraphics[width=1.98cm,height=1.3cm]{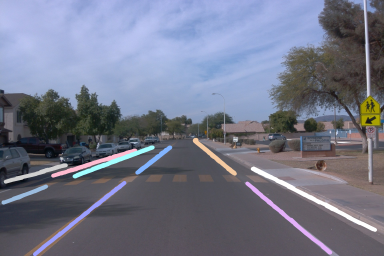}}\\[-1.98ex]

    \setcounter{subfigure}{-1}
    \subfloat {\raisebox{1.3em}{\rotatebox[origin=t]{90}{\scriptsize OpenLane-V}}}\hspace{-0.01cm}\,\!
    \subfloat {\includegraphics[width=1.98cm,height=1.3cm]{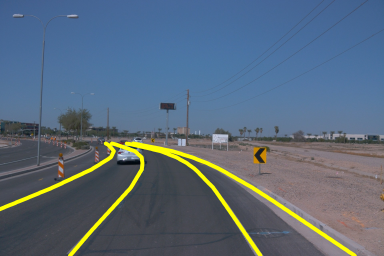}}\,\!\!
    \subfloat {\includegraphics[width=1.98cm,height=1.3cm]{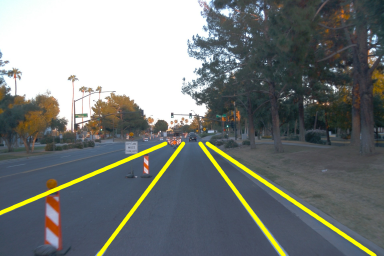}}\,\!\!
    \subfloat {\includegraphics[width=1.98cm,height=1.3cm]{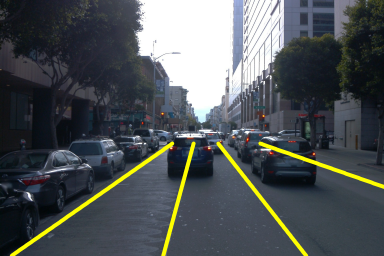}}\,\!\!
    \subfloat {\includegraphics[width=1.98cm,height=1.3cm]{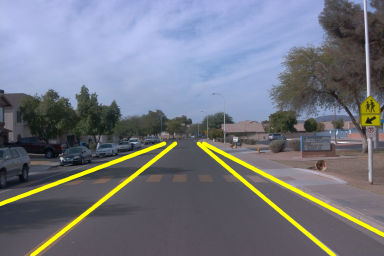}}

    \setcounter{subfigure}{-1}

    \caption
    {
        In OpenLane \cite{chen2022}, only visible lane parts are annotated, so the same lane may be split into multiple parts. In contrast, in OpenLane-V, each lane is seamlessly annotated, and invisible parts are inpainted. It is recommended to watch the accompanying video to see the improved annotations.
    }
    \label{fig:openlanev}
    \end{flushright}
    \vspace*{-0.3cm}
\end{figure}
\captionsetup[subfigure]{labelformat=parens}

\section{Datasets}\label{sec:dataset_modify}

\subsection{OpenLane-V}

OpenLane \cite{chen2022} is one of the largest lane datasets with about 200K images from 1,000 videos and about 880K lane annotations. Whereas both visible and invisible lanes are annotated in other datasets \cite{pan2018,tusimple,zhang2021}, OpenLane annotates only visible lane parts as in Figure~\ref{fig:openlanev}, which makes it unsuitable for video lane detection. Note that some parts are separated, but they actually belong to the same lane. Also, in challenging videos, lane parts are missing unpredictably, so they are not temporally consistent. We improve the annotations in OpenLane and construct a modified dataset, called OpenLane-V, by filling in missing parts semi-automatically based on matrix completion \cite{candes2010power}.

\begin{figure}[t]
  \centering
  \includegraphics[width=1\linewidth]{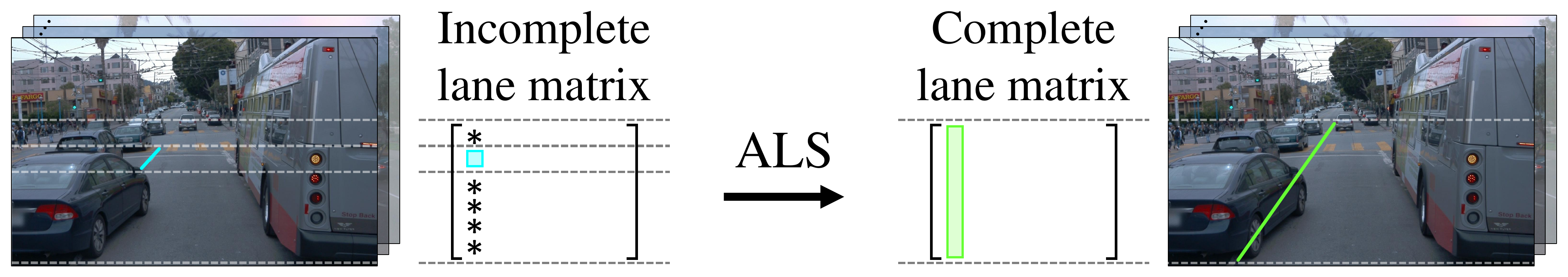}
  \caption{In OpenLane, only visible lane parts, such as the one in cyan, are annotated. Thus, the lane matrix is incomplete. Through ALS \cite{hastie2015}, we complete the matrix, so the missing parts are reconstructed and the whole lane in green is properly annotated.}
  \label{fig:ALS}
  \vspace*{-0.1cm}
\end{figure}

Figure~\ref{fig:ALS} illustrates the completion process. Let $\mathbf{A}=[\bfx_1, \bfx_2, \ldots, \bfx_L] \in \mathbb{R}^{N \times L}$ be a lane matrix \cite{jin2022} containing $L$ lanes in a dataset, where each lane $\bfx_i=[x_1, x_2, \ldots, x_N]^\top$ is represented by the $x$-coordinates of $N$ points sampled uniformly in the vertical direction. In OpenLane, however, some entries in $\mathbf{A}$ are not measured, since only visible lane parts are annotated. In other words, $\mathbf{A}$ is an incomplete matrix.

To complete $\mathbf{A}$, we perform factorization $\mathbf{A} \approx \mathbf{U} ^\top\mathbf{V}$, where $\mathbf{U} \in  \mathbb{R}^{R \times N}$, $\mathbf{V} \in \mathbb{R}^{R \times L}$, and $R$ is the rank of the factored $\mathbf{A}$. In \cite{jin2022}, it was shown that a lane matrix has low-rank characteristics, so we set $R=3$. Then, we find optimal $\mathbf{U}$ and $\mathbf{V}$ to minimize an objective function:
\begin{equation}\label{eq:objective}
    \min_{\bfU, \bfV} \sum_{(i, j) \in {\cal O}} (\mathbf{A}_{i,j} - \bfu^\top_{i} \bfv_j)^2 + \lambda (\sum_{i} {\| \bfu_i \|}^2 + \sum_{j} {\| \bfv_i \|}^2),
\end{equation}
where ${\cal O}$ is the coordinate set of observed entries, and $\lambda$ is a regularization parameter. Since finding the optimal factors is NP-hard \cite{hastie2015}, we adopt the alternating least square (ALS) algorithm \cite{hastie2015, jain2013}, in which we optimize $\bfV$ after fixing $\bfU$, and vice versa. This is repeated until convergence. The objective function becomes convex when one of the factors is fixed, so ALS is guaranteed to converge to a local minimum.

Entirely invisible lanes with no annotations, however, cannot be restored automatically. They disappear suddenly in videos, causing flickering artifacts. We manually annotate such lanes using the information in neighboring frames. Moreover, in some videos, too large a portion of lane parts is invisible, making the completed results unreliable. We remove such videos from OpenLane-V. This semi-automatic process is described in detail in the supplement (Section B).

As a result, OpenLane-V consists of about 90K images from 590 videos. It is split into a training set of 70K images from 450 videos and a test set of 20K images from 140 videos. As in the CULane dataset \cite{pan2018}, we annotate up to 4 road lanes in each image, which are ego and alternative lanes, to focus on important lanes for driving.

\subsection{VIL-100}
VIL-100 \cite{zhang2021} is the first dataset for video lane detection containing 100 videos. It is split into 80 training and 20 test videos. Each video has 100 frames. VIL-100 is less challenging than OpenLane-V and contains only 10K images.

\section{Experimental Results}

In addition to the results in this section, it is recommended to watch the accompanying video clips to compare video lane detection results more clearly.

\begin{figure}[t]
  \centering
  \includegraphics[width=1\linewidth]{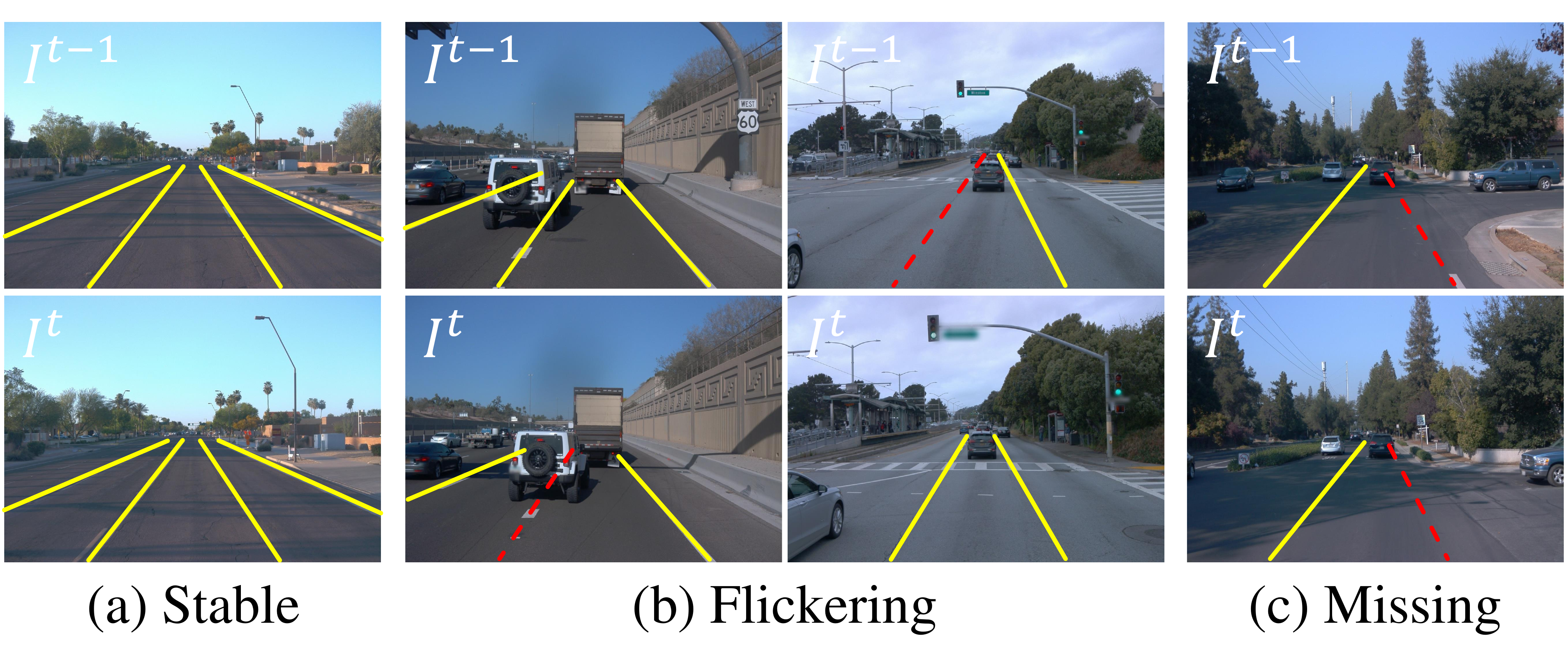}
  \caption{Three cases of lane detection results at consecutive frames: Correctly detected lanes are in yellow, while falsely dismissed ones are in red. In (a), each lane is detected correctly and stably in both frames. In (b), a detected lane at one frame is not detected at the other, leading to a flicker. In (c), the right lane is missed at both frames because it is not marked on the crossroad.}
  \label{fig:video_metric}
\end{figure}

\captionsetup[subfigure]{labelformat=empty}
\begin{figure*}[t]
\vspace{-0.3cm}

    \begin{flushright}
    \subfloat {\raisebox{1.2em}{\rotatebox[origin=t]{90}{\notsotiny Image}}}\hspace{-0.01cm}\,\!
    \subfloat {\includegraphics[width=2.12cm,height=1.15cm]{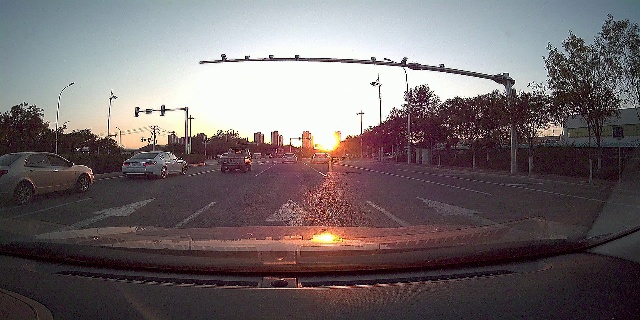}}\,\!\!
    \subfloat {\includegraphics[width=2.12cm,height=1.15cm]{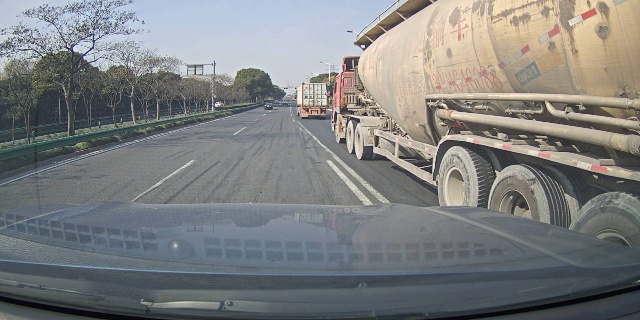}}\,\!\!
    \subfloat {\includegraphics[width=2.12cm,height=1.15cm]{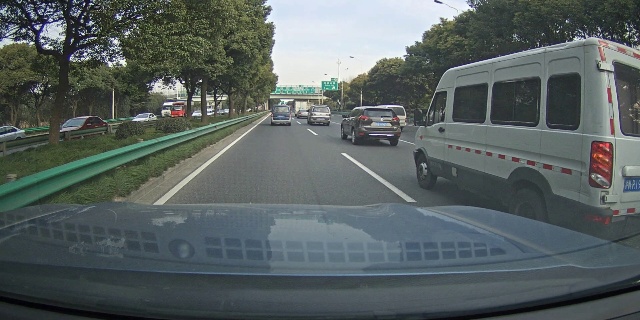}}\,\!\!
    \subfloat {\includegraphics[width=2.12cm,height=1.15cm]{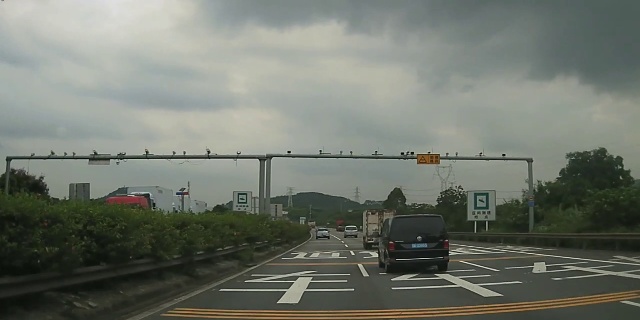}}\,\!\!
    \subfloat {\includegraphics[width=2.12cm,height=1.15cm]{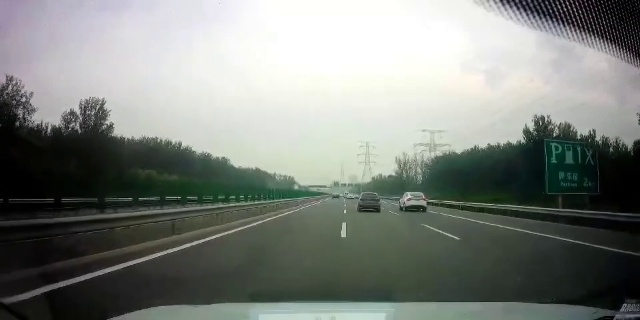}}\,\!\!
    \subfloat {\includegraphics[width=2.12cm,height=1.15cm]{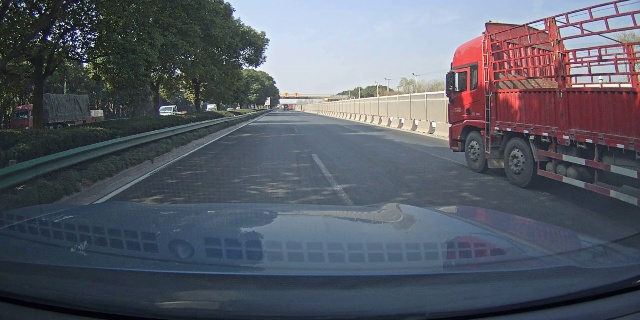}}\,\!\!
    \subfloat {\includegraphics[width=2.12cm,height=1.15cm]{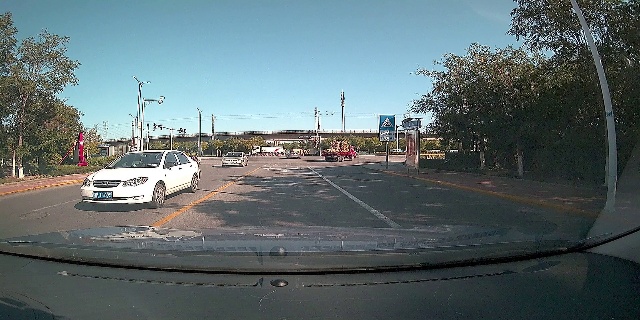}}\,\!\!
    \subfloat {\includegraphics[width=2.12cm,height=1.15cm]{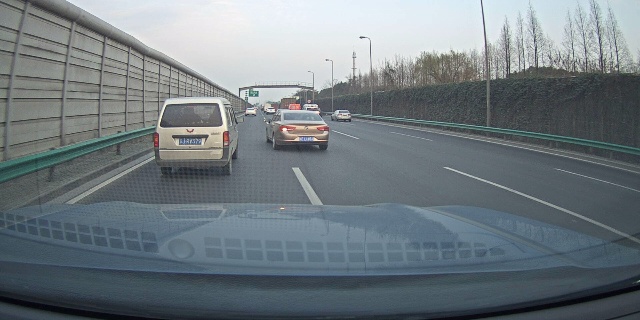}}\\[-2.2ex]

    \subfloat {\raisebox{1.2em}{\rotatebox[origin=t]{90}{\notsotiny MMA-Net}}}\hspace{-0.01cm}\,
    \subfloat {\includegraphics[width=2.12cm,height=1.15cm]{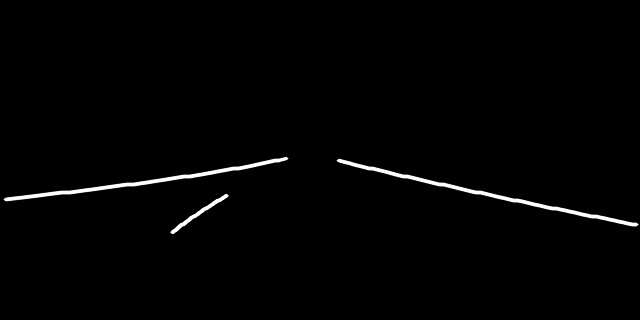}}\,\!\!
    \subfloat {\includegraphics[width=2.12cm,height=1.15cm]{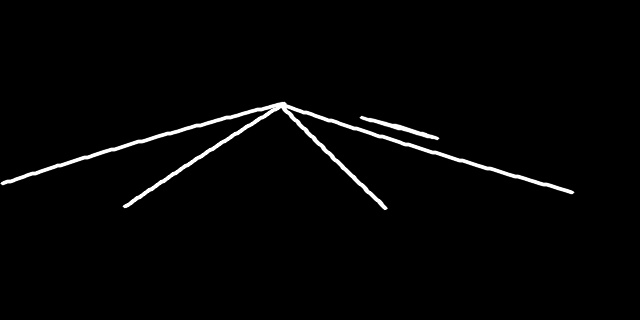}}\,\!\!
    \subfloat {\includegraphics[width=2.12cm,height=1.15cm]{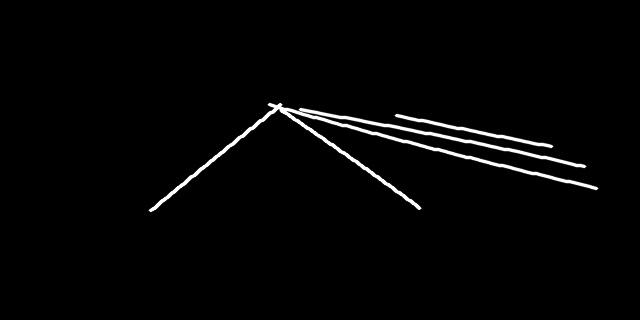}}\,\!\!
    \subfloat {\includegraphics[width=2.12cm,height=1.15cm]{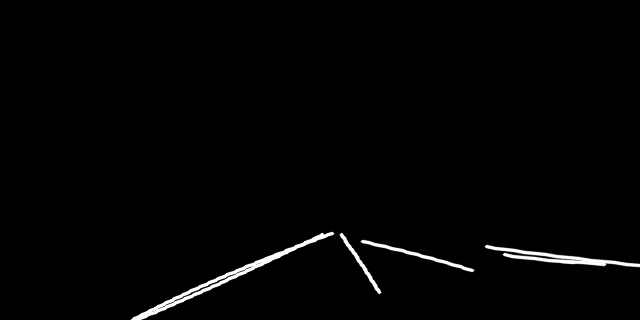}}\,\!\!
    \subfloat {\includegraphics[width=2.12cm,height=1.15cm]{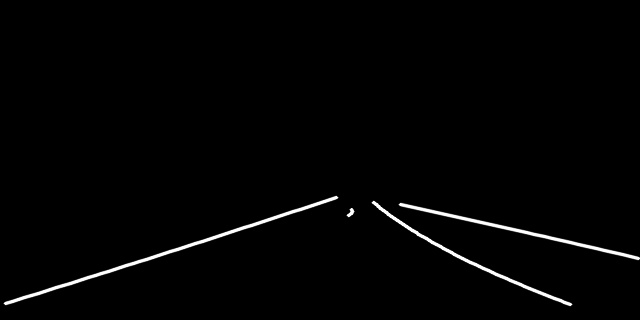}}\,\!\!
    \subfloat {\includegraphics[width=2.12cm,height=1.15cm]{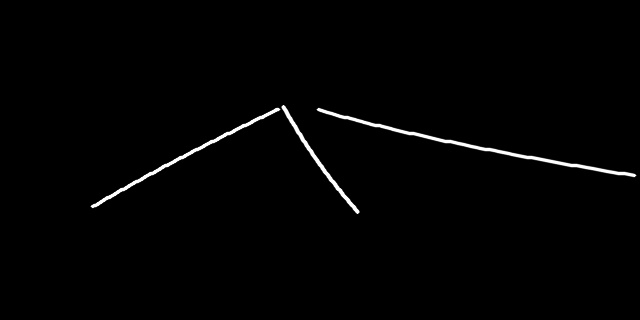}}\,\!\!
    \subfloat {\includegraphics[width=2.12cm,height=1.15cm]{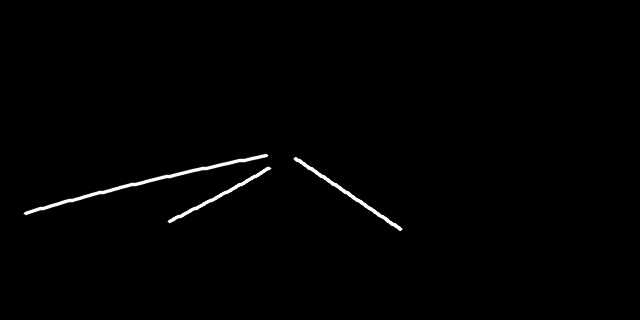}}\,\!\!
    \subfloat {\includegraphics[width=2.12cm,height=1.15cm]{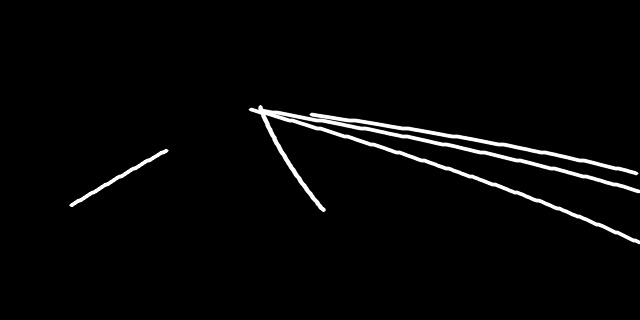}}\\[-2.2ex]

    \setcounter{subfigure}{-1}

    \subfloat {\raisebox{1.2em}{\rotatebox[origin=t]{90}{\notsotiny MFIALane}}}\hspace{-0.01cm}\,
    \subfloat {\includegraphics[width=2.12cm,height=1.15cm]{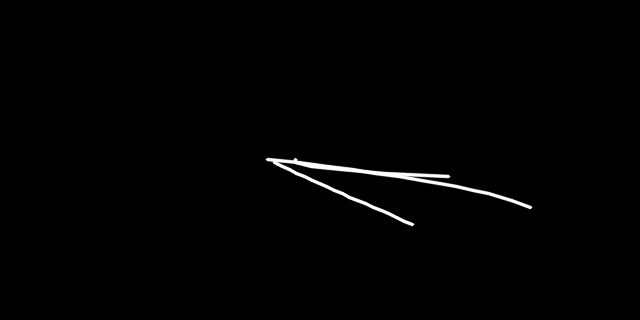}}\,\!\!
    \subfloat {\includegraphics[width=2.12cm,height=1.15cm]{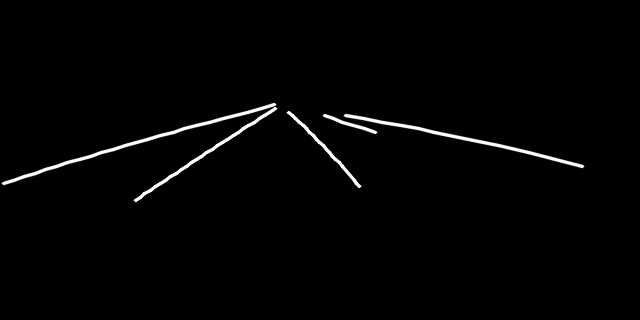}}\,\!\!
    \subfloat {\includegraphics[width=2.12cm,height=1.15cm]{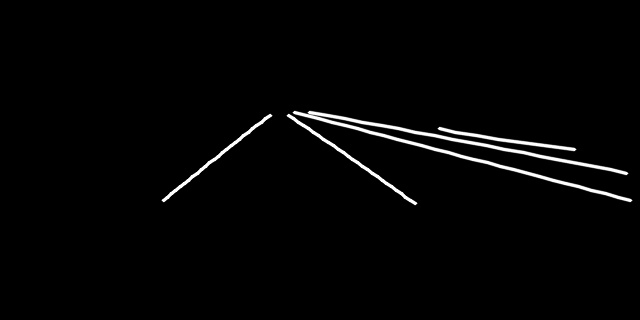}}\,\!\!
    \subfloat {\includegraphics[width=2.12cm,height=1.15cm]{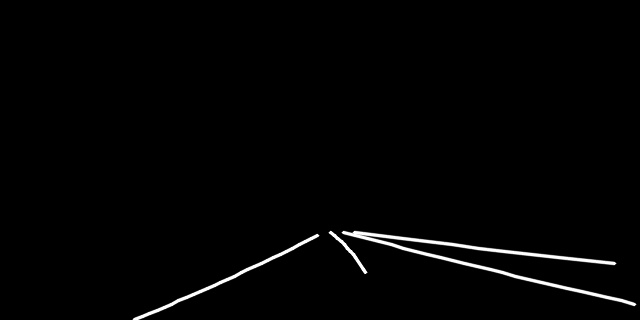}}\,\!\!
    \subfloat {\includegraphics[width=2.12cm,height=1.15cm]{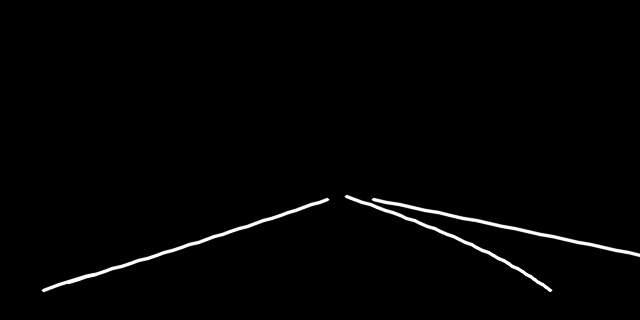}}\,\!\!
    \subfloat {\includegraphics[width=2.12cm,height=1.15cm]{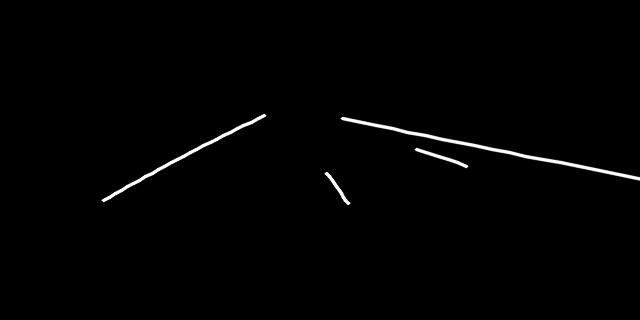}}\,\!\!
    \subfloat {\includegraphics[width=2.12cm,height=1.15cm]{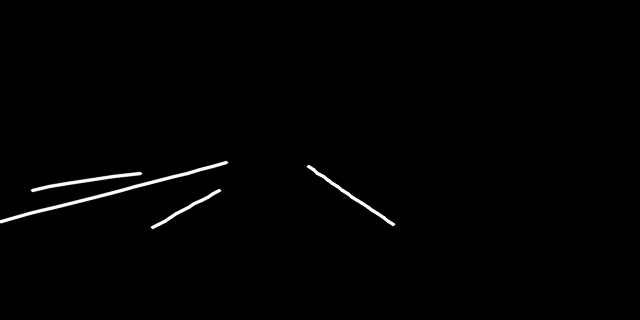}}\,\!\!
    \subfloat {\includegraphics[width=2.12cm,height=1.15cm]{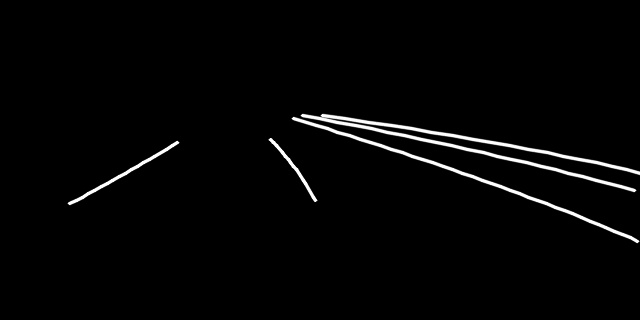}}\\[-2.2ex]

    \subfloat {\raisebox{1.2em}{\rotatebox[origin=t]{90}{\notsotiny RVLD}}}\hspace{-0.01cm}\,
    \subfloat {\includegraphics[width=2.12cm,height=1.15cm]{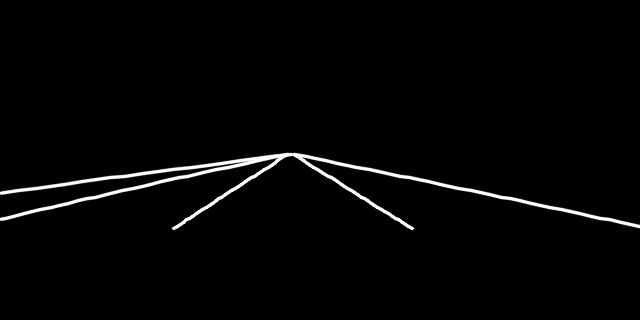}}\,\!\!
    \subfloat {\includegraphics[width=2.12cm,height=1.15cm]{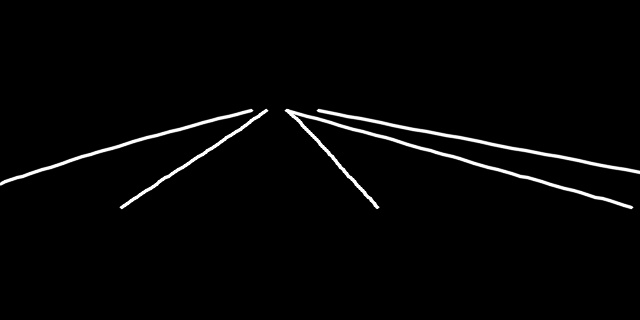}}\,\!\!
    \subfloat {\includegraphics[width=2.12cm,height=1.15cm]{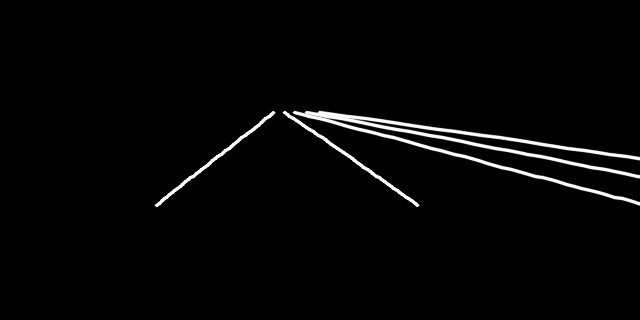}}\,\!\!
    \subfloat {\includegraphics[width=2.12cm,height=1.15cm]{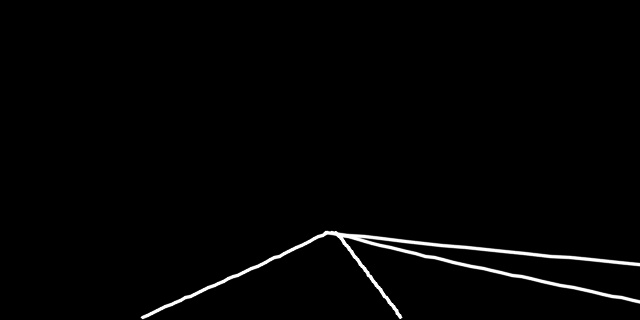}}\,\!\!
    \subfloat {\includegraphics[width=2.12cm,height=1.15cm]{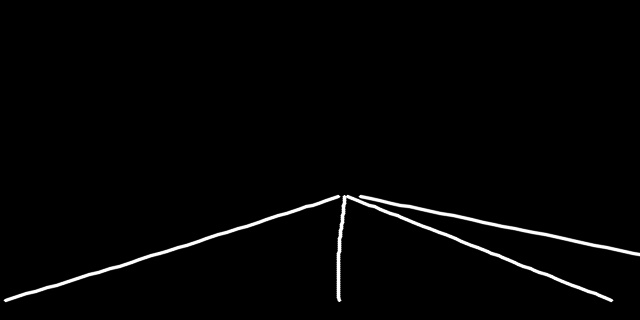}}\,\!\!
    \subfloat {\includegraphics[width=2.12cm,height=1.15cm]{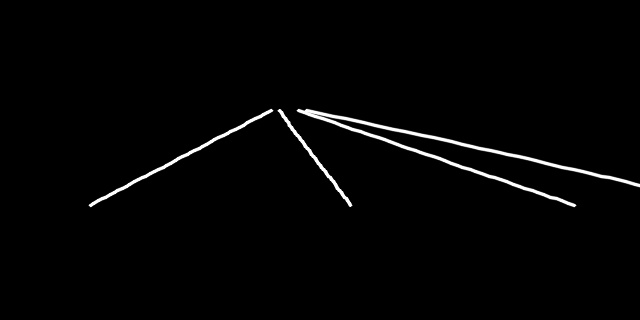}}\,\!\!
    \subfloat {\includegraphics[width=2.12cm,height=1.15cm]{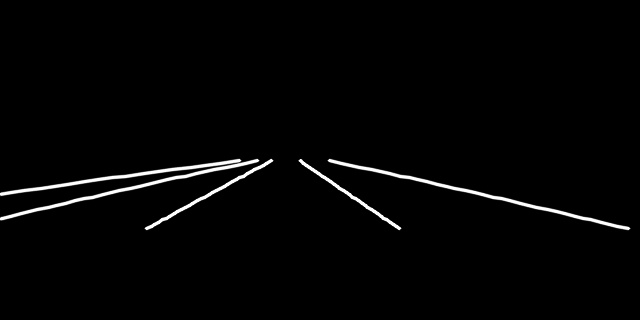}}\,\!\!
    \subfloat {\includegraphics[width=2.12cm,height=1.15cm]{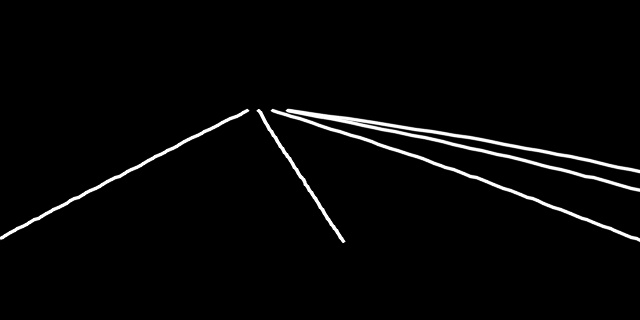}}\\[-2.2ex]

    \subfloat {\raisebox{1.2em}{\rotatebox[origin=t]{90}{\notsotiny Ground-truth}}}\hspace{-0.01cm}\,
    \subfloat {\includegraphics[width=2.12cm,height=1.15cm]{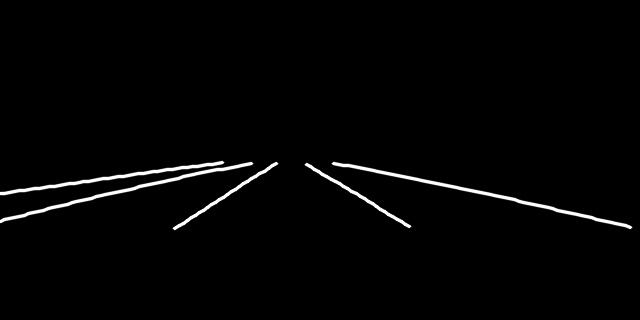}}\,\!\!
    \subfloat {\includegraphics[width=2.12cm,height=1.15cm]{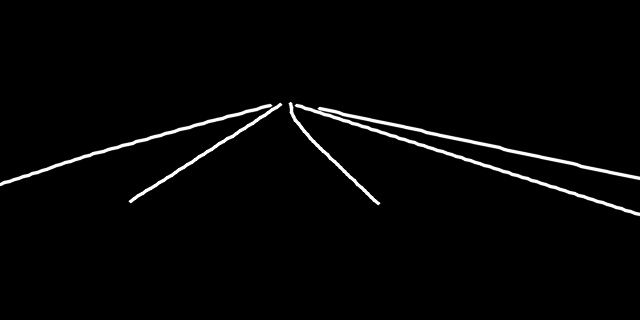}}\,\!\!
    \subfloat {\includegraphics[width=2.12cm,height=1.15cm]{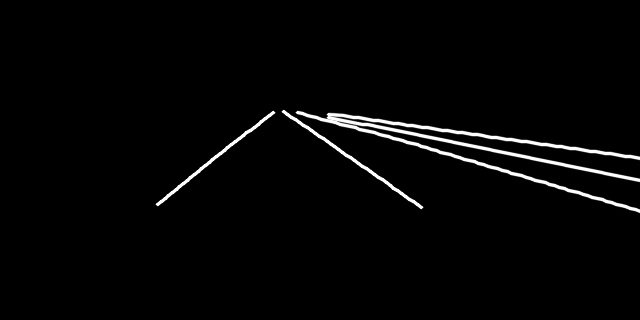}}\,\!\!
    \subfloat {\includegraphics[width=2.12cm,height=1.15cm]{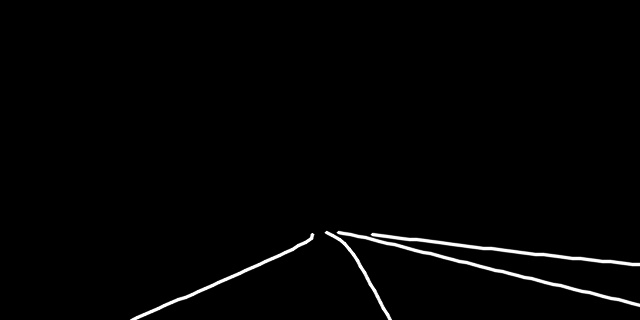}}\,\!\!
    \subfloat {\includegraphics[width=2.12cm,height=1.15cm]{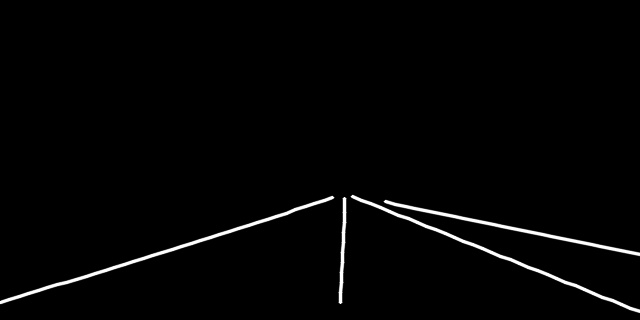}}\,\!\!
    \subfloat {\includegraphics[width=2.12cm,height=1.15cm]{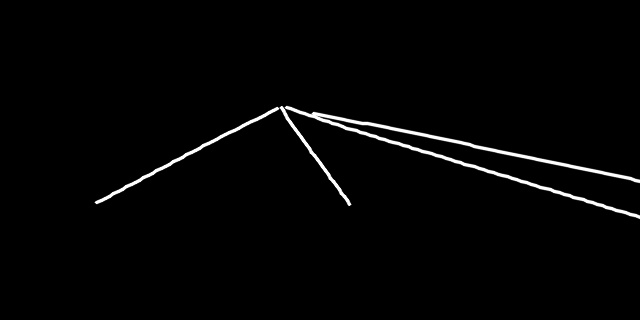}}\,\!\!
    \subfloat {\includegraphics[width=2.12cm,height=1.15cm]{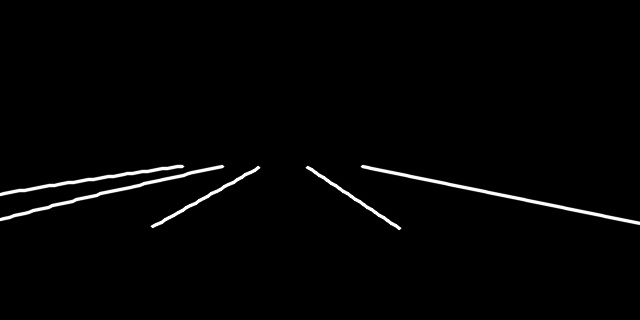}}\,\!\!
    \subfloat {\includegraphics[width=2.12cm,height=1.15cm]{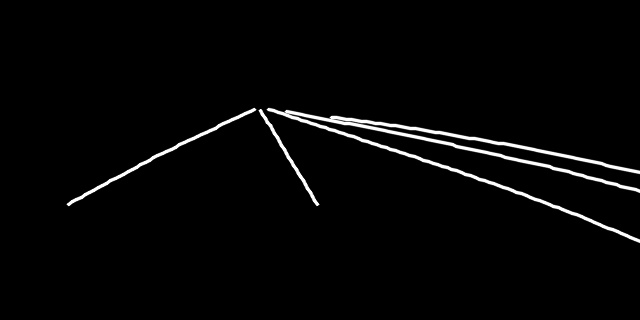}}
    \setcounter{subfigure}{-1}

    \caption
    {
        Comparison of lane detection results on the VIL-100 dataset.
    }
    \label{fig:vil100_result}
    \end{flushright}
    \vspace*{-0.3cm}
\end{figure*}
\captionsetup[subfigure]{labelformat=parens}

\subsection{Evaluation Metrics}

\noindent\textbf{Conventional metrics:}
For lane detection, image-based metrics are generally employed. Each lane is regarded as a thin stripe with 30 pixel width \cite{pan2018}. Then, a predicted lane is declared correct if its IoU ratio with GT is greater than $\tau$. The precision and the recall are computed by
\begin{equation}\label{eq:pre_rec}
    \textstyle
    {\rm Precision} = \frac{\rm TP}{\rm TP + FP}, \;\;\; {\rm Recall} = \frac{\rm TP}{\rm TP + FN}
\end{equation}
where $\rm TP$ is the number of correctly detected lanes, $\rm FP$ is that of false positives, and $\rm FN$ is that of false negatives. Then, the F-measure at threshold $\tau$ is defined as
\begin{equation}\label{eq:fscore}
    \textstyle
    {\rm F1}^{\tau} =  \frac{2 \times \rm Precision \times \rm Recall}{\rm Precision + \rm Recall}.
\end{equation}
Also, mIoU is computed by averaging the IoU scores of correctly detected lanes.

\vspace*{0.1cm}
\noindent\textbf{Video metrics:}
We propose two video metrics to assess the temporal stability of detected lanes. When a lane suddenly disappears or a new lane is suddenly detected at a frame, vehicle maneuvers are hard to control, possibly leading to dangerous situations. It is important to achieve temporally stable lane detection. There are three cases for a matching pair of lanes at adjacent frames: \textit{Stable}, \textit{Flickering}, and \textit{Missing}. A stable case is one where a lane is successfully detected at both frames, as shown in Figure~\ref{fig:video_metric}(a). In a flickering case, a lane is detected at one frame but missed at the other, as in Figure~\ref{fig:video_metric}(b). A missing case is the worst one in which a lane is missed consecutively at both frames, as in Figure~\ref{fig:video_metric}(c).

Let $\rm N$ be the number of GT lanes that have matching instances at previous frames, and let $\rm N_S$, $\rm N_F$, $\rm N_M$ be the numbers of stable, flickering, and missing cases, respectively. Note that $\rm N= N_S +  N_F +  N_M$. Then, we define the flickering and missing rates as
\begin{equation}\label{eq:pre_rec}
    \textstyle
    \rm
    {\rm R}_{\rm F}^\tau = \frac{N_F}{N}, \quad {\rm R}_{\rm M}^\tau = \frac{N_M}{N},
\end{equation}
where $\tau$ is the IoU threshold for correct detection.

\begin{table}[t]\centering
    \renewcommand{\arraystretch}{0.9}
    \caption
    {
        Comparison of mIoU and $\rm F1$ scores on VIL-100: image lane detectors and video ones are listed separately.
    }
    \vspace*{-0.15cm}
    \resizebox{0.9\linewidth}{!}{
    \begin{tabular}[t]{+L{3.0cm}^C{1.6cm}^C{1.6cm}^C{1.6cm}}
    \toprule
    & mIoU & ${\rm F1}^{0.5}$ & ${\rm F1}^{0.8}$\\
    \midrule
         LaneNet \cite{neven2018}        & 0.633 & 0.721 & 0.222 \\
         ENet-SAD \cite{hou2019road}     & 0.616 & 0.755 & 0.205 \\
         LSTR \cite{liu2021}             & 0.573 & 0.703 & 0.131 \\
         RESA \cite{zheng2021}           & 0.702 & 0.874 & 0.345 \\
         LaneATT \cite{tabelini2021CVPR}     & 0.664 & 0.823 & - \\
         MFIALane \cite{qiu2022mfialane} & - & \underline{0.905} & \underline{0.565}\\
    \midrule
        MMA-Net \cite{zhang2021}         & 0.705 & 0.839 & 0.458 \\
        LaneATT-T \cite{tabelini2022}    & 0.692 & 0.846 & - \\
        TGC-Net \cite{wang2022video}     & \underline{0.738} & 0.892 & 0.469 \\
    \midrule
         RVLD {\scriptsize (Proposed)}                       & \bf{0.787} & \bf{0.924} & \bf{0.582}\\
    \bottomrule
    \end{tabular}}
    \label{table:vil100img}
\end{table}

\begin{table}[t]\centering
    \renewcommand{\arraystretch}{0.9}
    \caption
    {
        Flickering and missing rates on VIL-100.
    }
    \vspace*{-0.15cm}
    \resizebox{0.9\linewidth}{!}{
    \begin{tabular}[t]{+L{2.8cm}^C{1.2cm}^C{1.2cm}^C{1.2cm}^C{1.2cm}}
    \toprule
    & ${\rm R}_{\rm F}^{0.5}$ & ${\rm R}_{\rm M}^{0.5}$ & ${\rm R}_{\rm F}^{0.8}$ & ${\rm R}_{\rm M}^{0.8}$\\
    \midrule

        MMA-Net \cite{zhang2021}         & 0.047 & 0.128 & \bf{0.200} & 0.428\\
        MFIALane \cite{qiu2022mfialane} & \underline{0.042} & \underline{0.127} & 0.206 & \underline{0.323}\\
        RVLD {\scriptsize (Proposed)}                        & \bf{0.038} & \bf{0.050} & \underline{0.203} & \bf{0.306}\\
    \bottomrule
    \end{tabular}}
    \label{table:vil100video}
\end{table}

\begin{table*}[t]\centering
    \vspace*{-0.2cm}
    \renewcommand{\arraystretch}{0.9}
    \caption
    {
        Comparison on OpenLane-V.
    }
    \vspace*{-0.15cm}
    \resizebox{0.9\linewidth}{!}{
    \begin{tabular}[t]{+L{3.8cm}^C{1.7cm}^C{1.7cm}^C{1.7cm}^C{1.7cm}^C{1.7cm}^C{1.7cm}^C{1.7cm}}
    \toprule
    & mIoU & ${\rm F1}^{0.5}$ & ${\rm F1}^{0.8}$ & ${\rm R}_{\rm F}^{0.5}$ & ${\rm R}_{\rm M}^{0.5}$ & ${\rm R}_{\rm F}^{0.8}$ & ${\rm R}_{\rm M}^{0.8}$\\
    \midrule
         ConvLSTM \cite{zou2019robust}               & 0.529 & 0.641 & 0.353 & 0.058 & 0.282 & 0.091 & 0.574 \\
         ConvGRUs \cite{zhang2021lane}               & 0.540 & 0.641 & 0.355 & 0.064 & 0.288 & 0.094 & 0.576 \\
         MMA-Net \cite{zhang2021}               & 0.574 & 0.573 & 0.328 & \underline{0.044} & 0.461 & \underline{0.071} & 0.671 \\
         MFIALane \cite{qiu2022mfialane}        & 0.697 & 0.723 & 0.475 & 0.061 & 0.300 & 0.080 & 0.519\\
         CondLaneNet \cite{liu2021condlanenet}  & 0.698 & 0.780 & 0.450 & 0.047 & 0.239 & 0.084 & 0.531\\
         GANet \cite{wang2022}                  & 0.716 & \underline{0.801} & 0.530 & 0.048 & \underline{0.198} & 0.082 & 0.443\\
         CLRNet \cite{zheng2022}                & \bf{0.735} & 0.789 & \underline{0.554} & 0.054 & 0.224 & 0.086 & \underline{0.430}\\
    \midrule
         RVLD {\scriptsize (Proposed)}                   & \underline{0.727} & \bf{0.825} & \bf{0.566} & \bf{0.014} & \bf{0.167} & \bf{0.051} & \bf{0.406} \\
    \bottomrule
    \end{tabular}}
    \label{table:openlane}
    \vspace*{-0.1cm}
\end{table*}

\subsection{Comparative Assessment}

\noindent\textbf{VIL-100:}
We compare the proposed RVLD with conventional image lane detectors \cite{neven2018,hou2019road,liu2021,zheng2021,tabelini2021CVPR,qiu2022mfialane} and video ones \cite{zhang2021,tabelini2022,wang2022video} on VIL-100.
Table~\ref{table:vil100img} lists the mIoU and $\rm F1$ scores. RVLD outperforms all conventional algorithms in all image metrics. Especially, RVLD is better than the state-of-the-art video lane detector TGC-Net by significant margins of 0.049, 0.032, and 0.113 in mIoU, ${\rm F1}^{0.5}$, and ${\rm F1}^{0.8}$, respectively. Note that RVLD uses a single previous frame only, whereas the existing video lane detectors \cite{zhang2021,tabelini2022,wang2022video} use two or more past frames as input. MFIALane, an image lane detector, yields high $\rm F1$ scores, but it underperforms as compared with RVLD. This is because it processes each image independently and may fail to detect implied lanes. In contrast, RVLD detects lanes in a current frame more reliably by exploiting past information.

Table~\ref{table:vil100video} compares the ${\rm R}_{\rm F}$ and ${\rm R}_{\rm M}$ rates. LaneATT-T and TGC-Net are not compared because their source codes are unavailable. MMA-Net achieves the lowest flickering rate ${\rm R}_{\rm F}^{0.8}$, but it does so because its missing rate ${\rm R}_{\rm M}^{0.8}$ is too high. Except for ${\rm R}_{\rm F}^{0.8}$, the proposed RVLD yields the lowest flickering and missing rates, indicating that RVLD provides temporally more stable detection results.

Figure~\ref{fig:vil100_result} presents some detection results. MMA-Net does not detect unobvious lanes precisely, even though it uses several past frames as input. MFIALane also fails to process those lanes reliably, for it is image-based. In contrast, the proposed RVLD provides better results using past information effectively.

\captionsetup[subfigure]{labelformat=empty}
\begin{figure*}[t]
\vspace{-0.2cm}

    \begin{flushright}
    \subfloat {\raisebox{1.2em}{\rotatebox[origin=t]{90}{\notsotiny Image}}}\hspace{-0.01cm}\,\!
    \subfloat {\includegraphics[width=2.12cm,height=1.15cm]{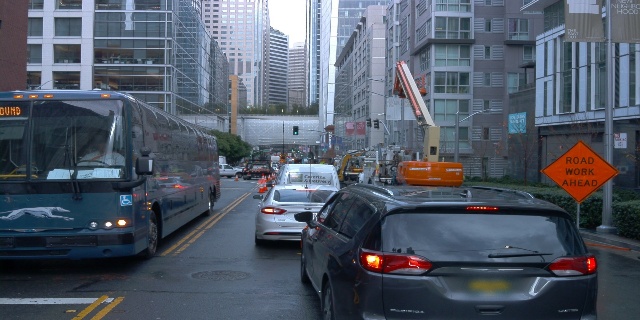}}\,\!\!
    \subfloat {\includegraphics[width=2.12cm,height=1.15cm]{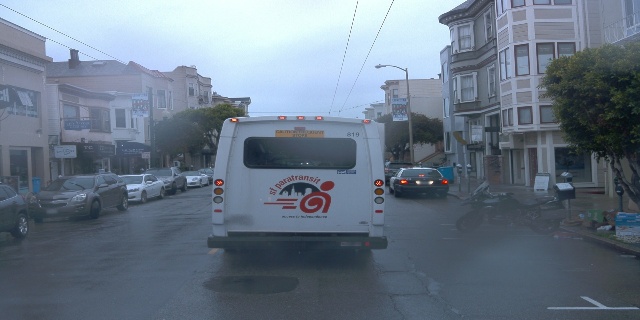}}\,\!\!
    \subfloat {\includegraphics[width=2.12cm,height=1.15cm]{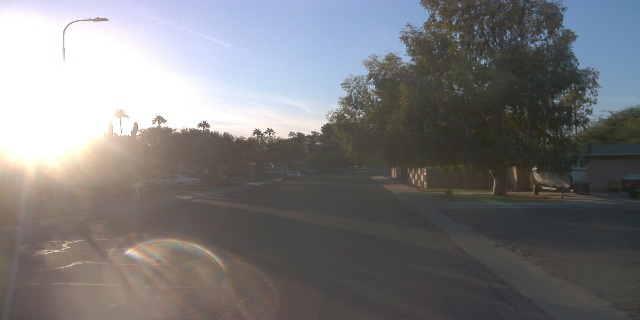}}\,\!\!
    \subfloat {\includegraphics[width=2.12cm,height=1.15cm]{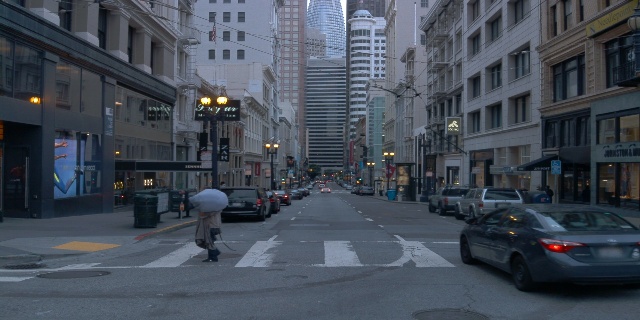}}\,\!\!
    \subfloat {\includegraphics[width=2.12cm,height=1.15cm]{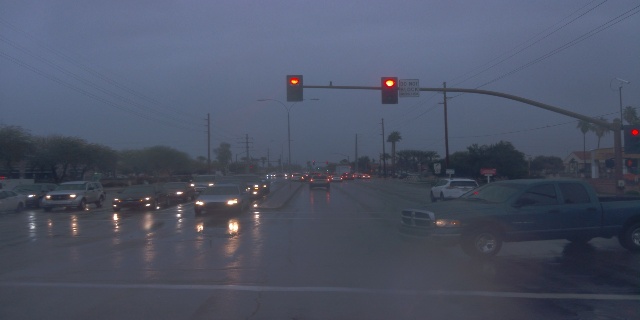}}\,\!\!
    \subfloat {\includegraphics[width=2.12cm,height=1.15cm]{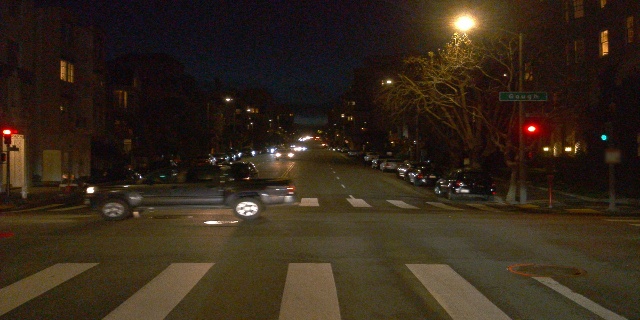}}\,\!\!
    \subfloat {\includegraphics[width=2.12cm,height=1.15cm]{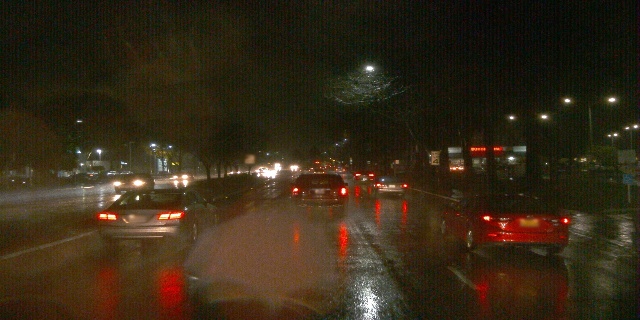}}\,\!\!
    \subfloat {\includegraphics[width=2.12cm,height=1.15cm]{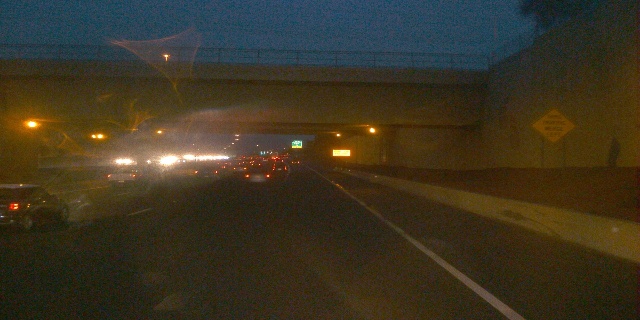}}\\[-2.2ex]

    \setcounter{subfigure}{-1}


    \subfloat {\raisebox{1.2em}{\rotatebox[origin=t]{90}{\notsotiny GANet}}}\hspace{-0.01cm}\,
    \subfloat {\includegraphics[width=2.12cm,height=1.15cm]{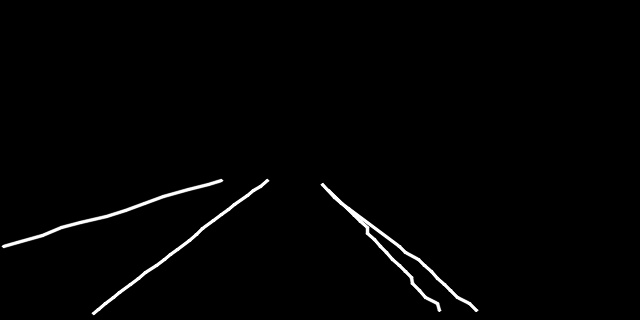}}\,\!\!
    \subfloat {\includegraphics[width=2.12cm,height=1.15cm]{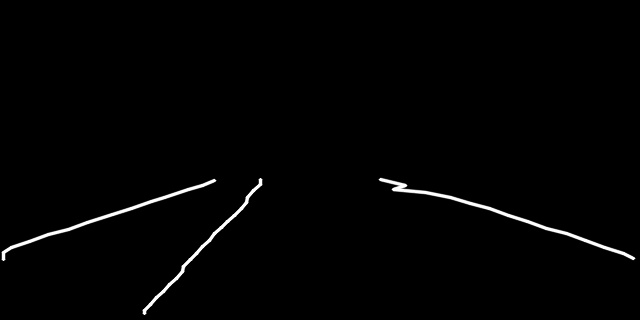}}\,\!\!
    \subfloat {\includegraphics[width=2.12cm,height=1.15cm]{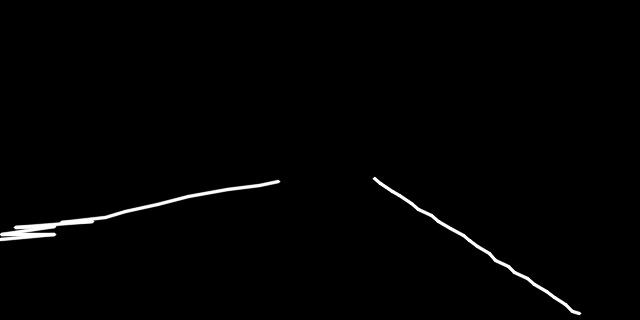}}\,\!\!
    \subfloat {\includegraphics[width=2.12cm,height=1.15cm]{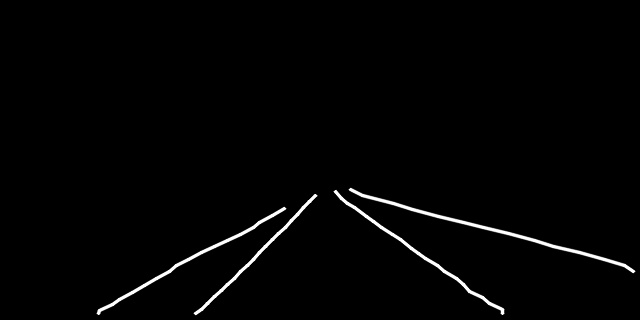}}\,\!\!
    \subfloat {\includegraphics[width=2.12cm,height=1.15cm]{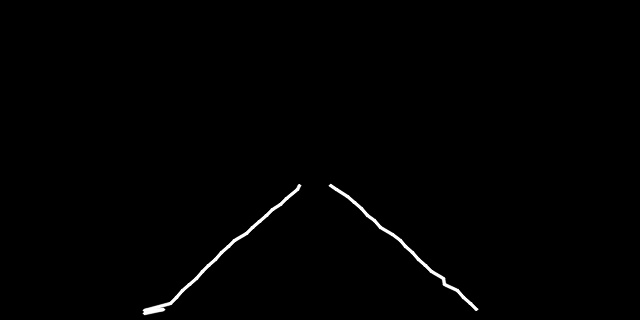}}\,\!\!
    \subfloat {\includegraphics[width=2.12cm,height=1.15cm]{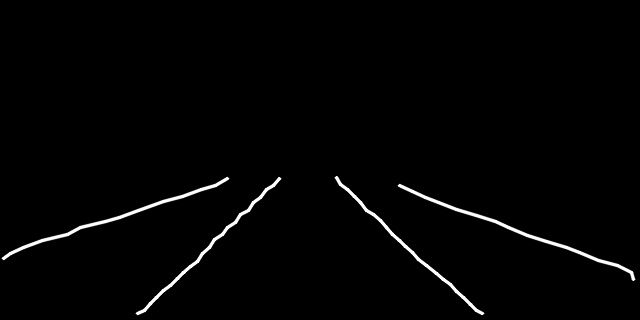}}\,\!\!
    \subfloat {\includegraphics[width=2.12cm,height=1.15cm]{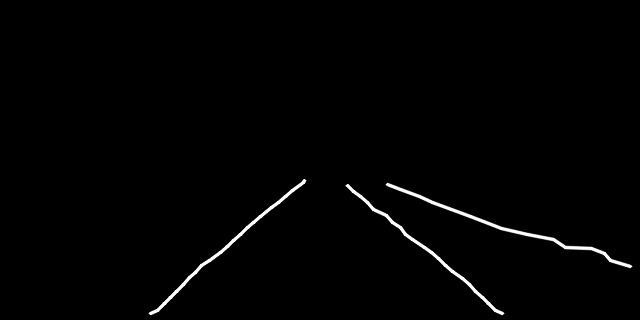}}\,\!\!
    \subfloat {\includegraphics[width=2.12cm,height=1.15cm]{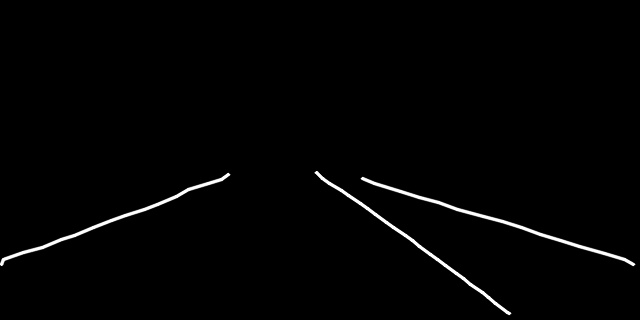}}\\[-2.2ex]

    \subfloat {\raisebox{1.2em}{\rotatebox[origin=t]{90}{\notsotiny CLRNet}}}\hspace{-0.01cm}\,
    \subfloat {\includegraphics[width=2.12cm,height=1.15cm]{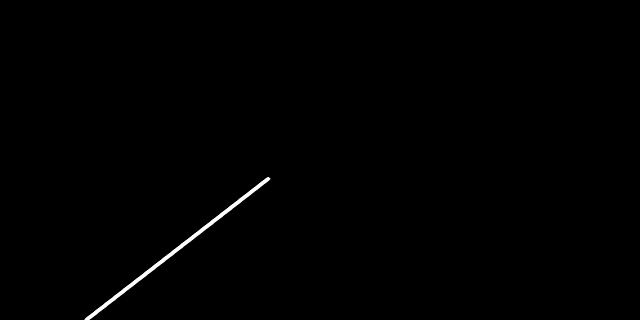}}\,\!\!
    \subfloat {\includegraphics[width=2.12cm,height=1.15cm]{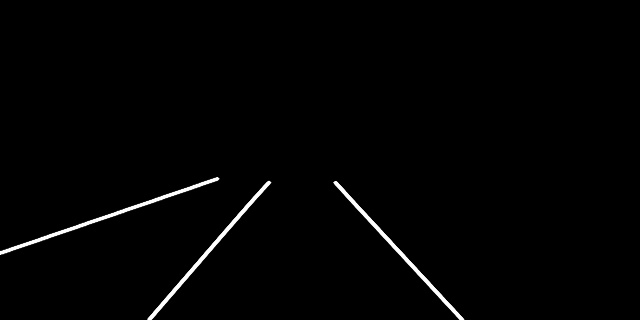}}\,\!\!
    \subfloat {\includegraphics[width=2.12cm,height=1.15cm]{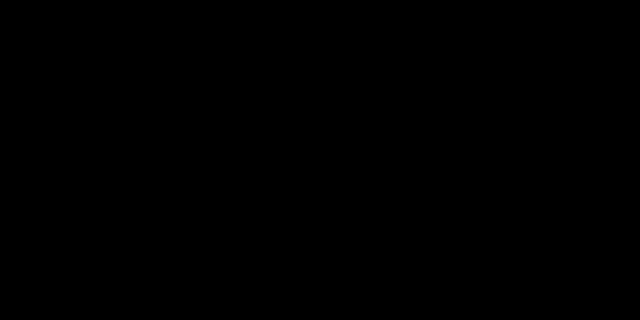}}\,\!\!
    \subfloat {\includegraphics[width=2.12cm,height=1.15cm]{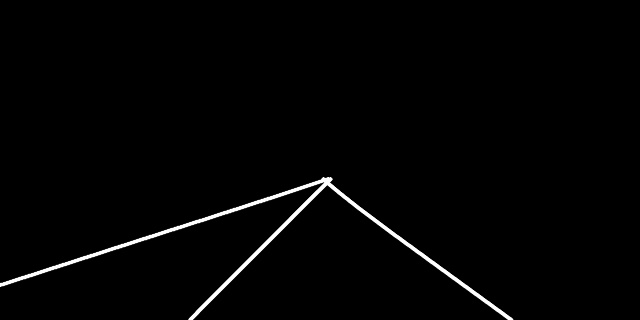}}\,\!\!
    \subfloat {\includegraphics[width=2.12cm,height=1.15cm]{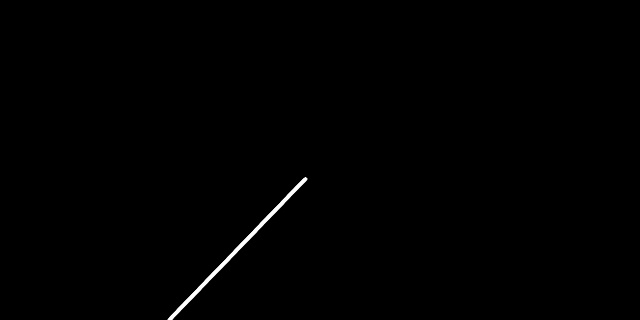}}\,\!\!
    \subfloat {\includegraphics[width=2.12cm,height=1.15cm]{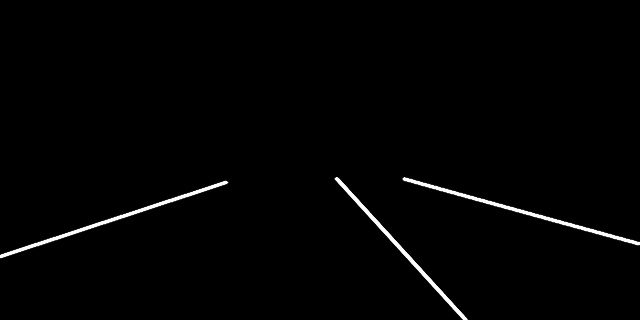}}\,\!\!
    \subfloat {\includegraphics[width=2.12cm,height=1.15cm]{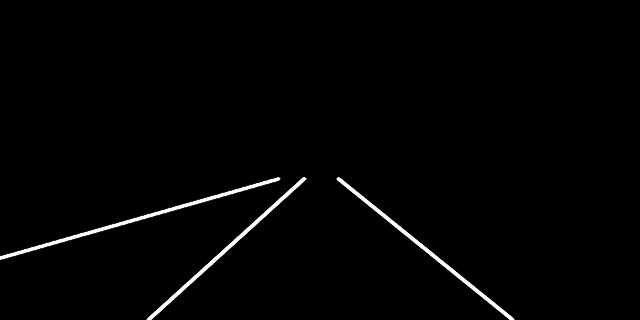}}\,\!\!
    \subfloat {\includegraphics[width=2.12cm,height=1.15cm]{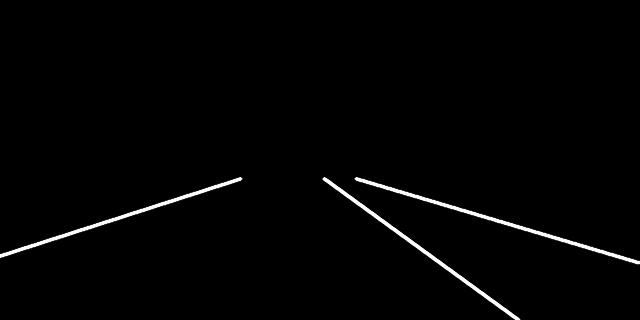}}\\[-2.2ex]

    \setcounter{subfigure}{-1}

    \subfloat {\raisebox{1.2em}{\rotatebox[origin=t]{90}{\notsotiny RVLD}}}\hspace{-0.01cm}\,
    \subfloat {\includegraphics[width=2.12cm,height=1.15cm]{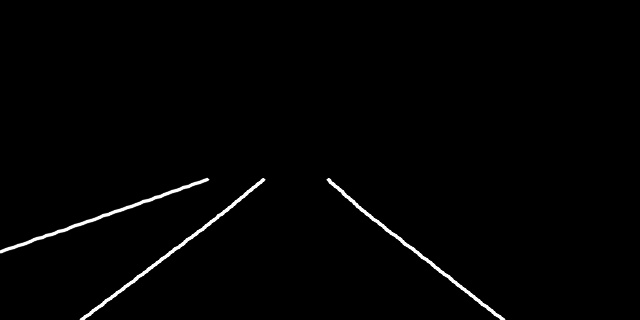}}\,\!\!
    \subfloat {\includegraphics[width=2.12cm,height=1.15cm]{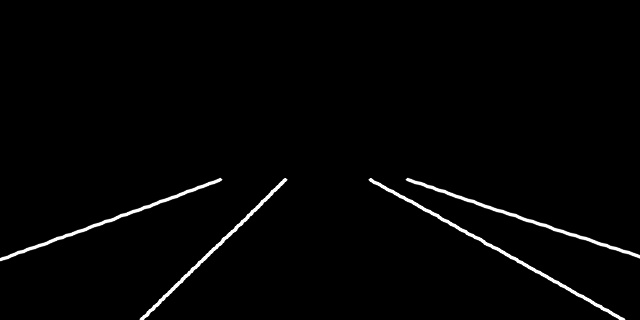}}\,\!\!
    \subfloat {\includegraphics[width=2.12cm,height=1.15cm]{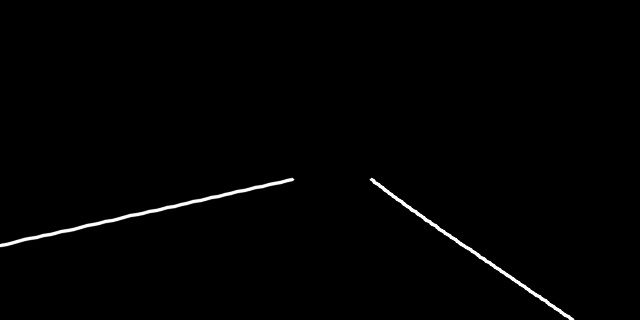}}\,\!\!
    \subfloat {\includegraphics[width=2.12cm,height=1.15cm]{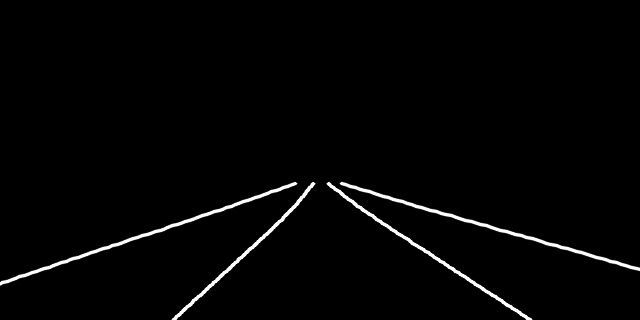}}\,\!\!
    \subfloat {\includegraphics[width=2.12cm,height=1.15cm]{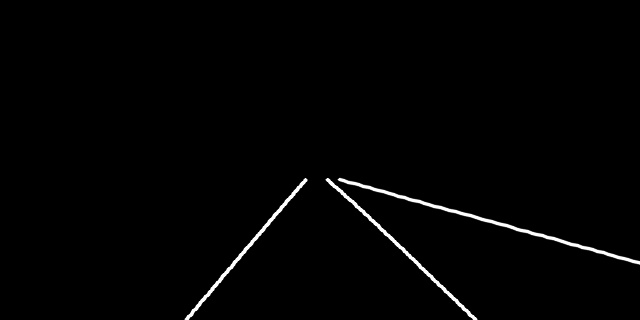}}\,\!\!
    \subfloat {\includegraphics[width=2.12cm,height=1.15cm]{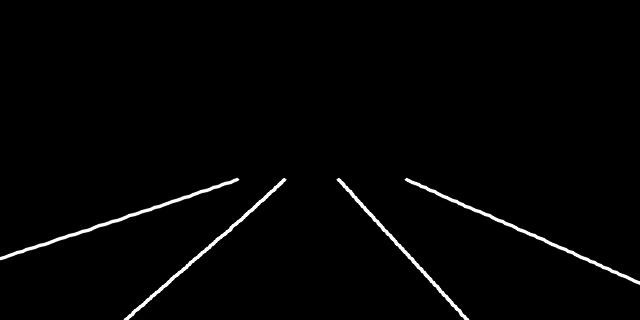}}\,\!\!
    \subfloat {\includegraphics[width=2.12cm,height=1.15cm]{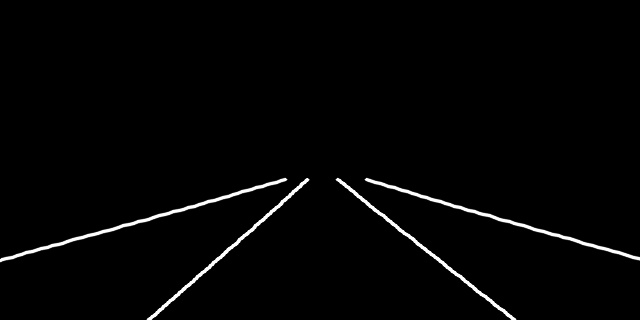}}\,\!\!
    \subfloat {\includegraphics[width=2.12cm,height=1.15cm]{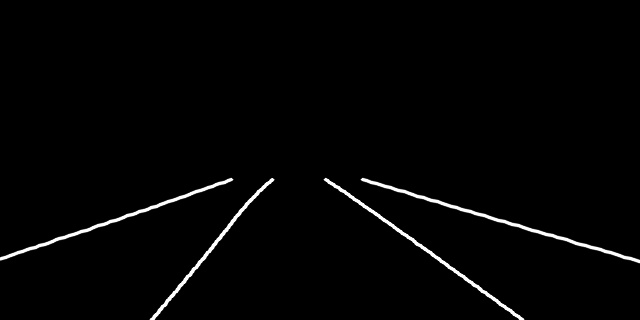}}\\[-2.2ex]

    \subfloat {\raisebox{1.2em}{\rotatebox[origin=t]{90}{\notsotiny Ground-truth}}}\hspace{-0.01cm}\,
    \subfloat {\includegraphics[width=2.12cm,height=1.15cm]{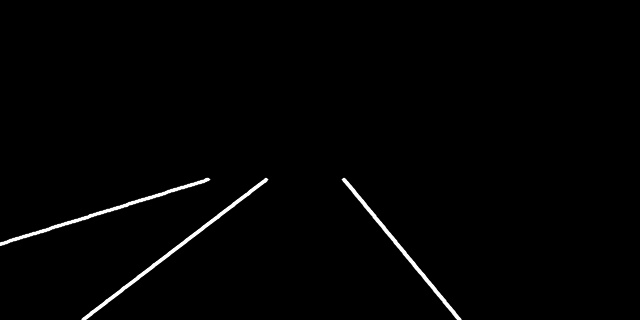}}\,\!\!
    \subfloat {\includegraphics[width=2.12cm,height=1.15cm]{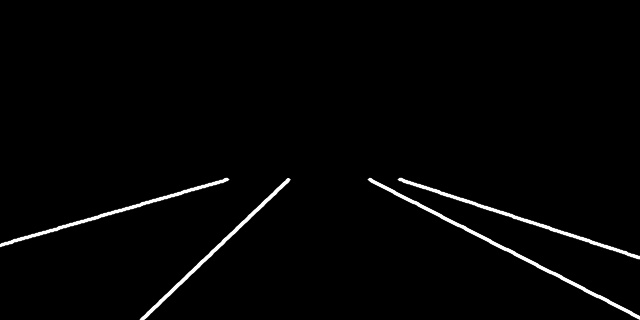}}\,\!\!
    \subfloat {\includegraphics[width=2.12cm,height=1.15cm]{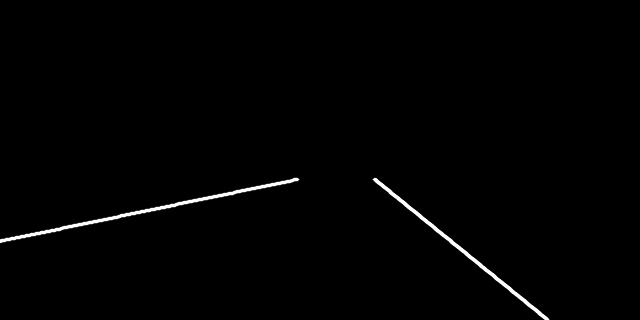}}\,\!\!
    \subfloat {\includegraphics[width=2.12cm,height=1.15cm]{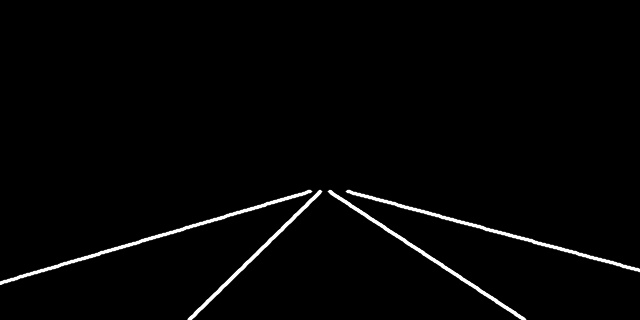}}\,\!\!
    \subfloat {\includegraphics[width=2.12cm,height=1.15cm]{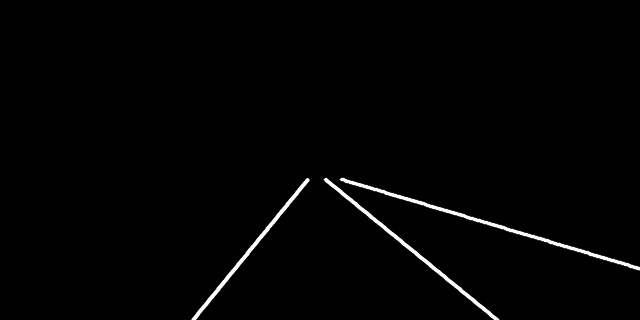}}\,\!\!
    \subfloat {\includegraphics[width=2.12cm,height=1.15cm]{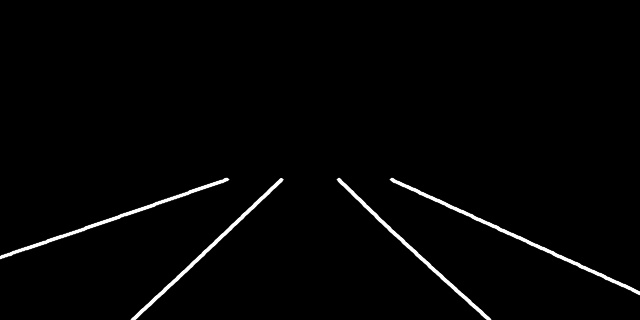}}\,\!\!
    \subfloat {\includegraphics[width=2.12cm,height=1.15cm]{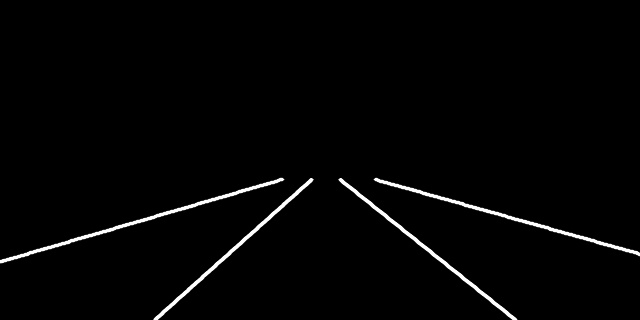}}\,\!\!
    \subfloat {\includegraphics[width=2.12cm,height=1.15cm]{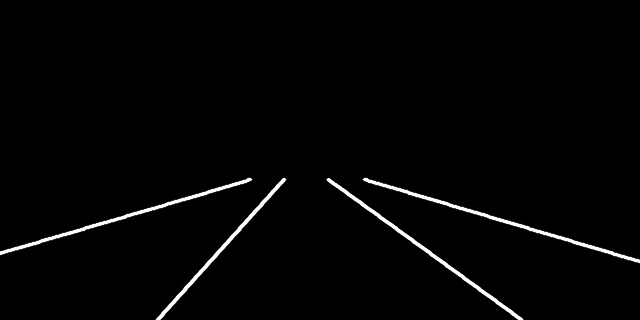}}
    \setcounter{subfigure}{-1}

    \caption
    {
        Comparison of lane detection results on the OpenLane-V dataset.
    }
    \label{fig:openlane_result}
    \end{flushright}
    \vspace*{-0.5cm}
\end{figure*}
\captionsetup[subfigure]{labelformat=parens}

\begin{figure}[t]
  \centering
  \includegraphics[width=1\linewidth]{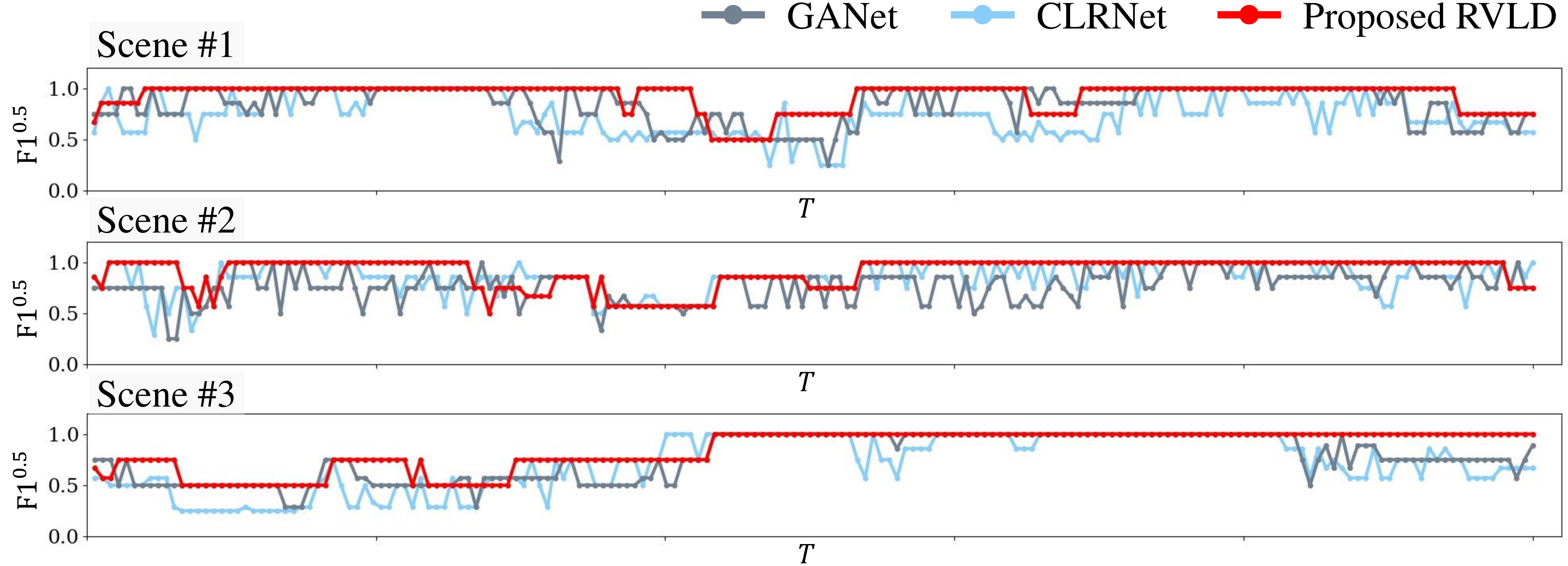}
  \caption{Comparison of ${\rm F1}^{0.5}$ scores per frame in three challenging videos in OpenLane-V. Compared to RVLD, GANet and CLRNet suffer from more fluctuating results. The frame-by-frame comparisons of lane detection results of CLRNet and RVLD on these three scenes are available in a supplemental video clip.}
  \vspace*{-0.2cm}
  \label{fig:graph}
\end{figure}

\captionsetup[subfigure]{labelformat=empty}
\begin{figure*}[t]
\vspace{-0.4cm}
    \begin{flushright}
    \subfloat {\raisebox{1.2em}{\rotatebox[origin=t]{90}{\notsotiny Image}}}\hspace{-0.01cm}\,\!
    \subfloat {\includegraphics[width=2.42cm,height=1.1cm]{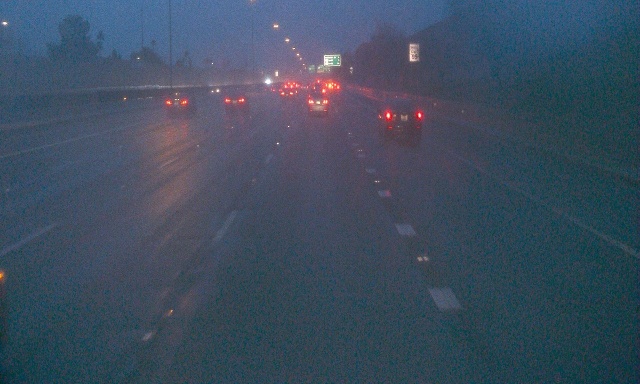}}\,\!\!
    \subfloat {\includegraphics[width=2.42cm,height=1.1cm]{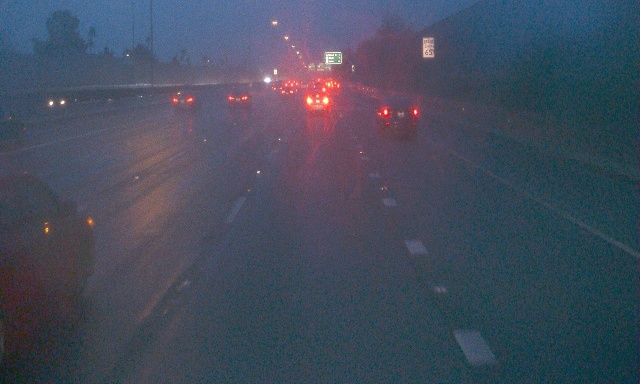}}\,\!\!
    \subfloat {\includegraphics[width=2.42cm,height=1.1cm]{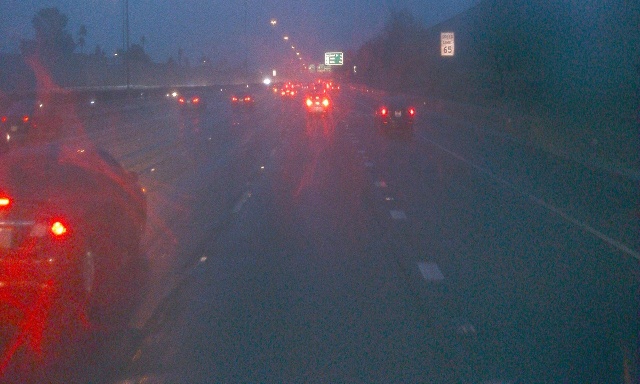}}\,\!\!
    \subfloat {\includegraphics[width=2.42cm,height=1.1cm]{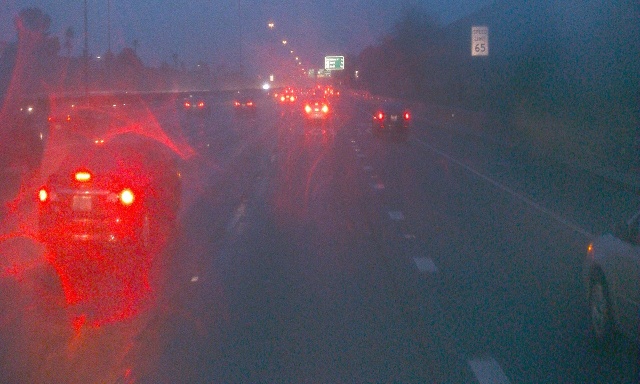}}\,\!\!
    \subfloat {\includegraphics[width=2.42cm,height=1.1cm]{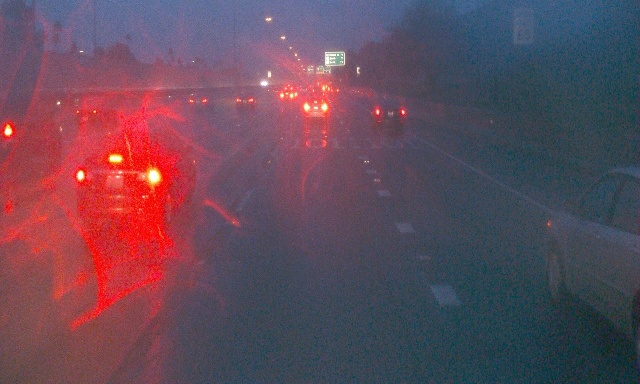}}\,\!\!
    \subfloat {\includegraphics[width=2.42cm,height=1.1cm]{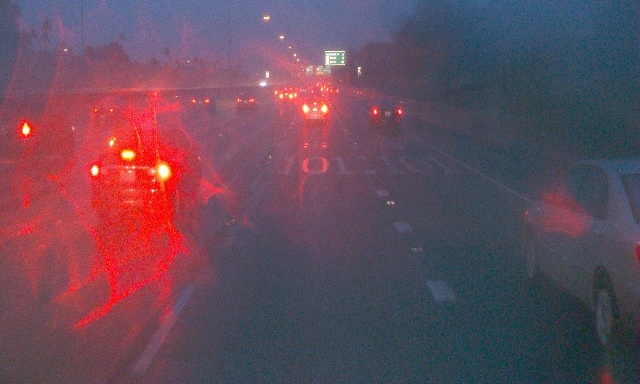}}\,\!\!
    \subfloat {\includegraphics[width=2.42cm,height=1.1cm]{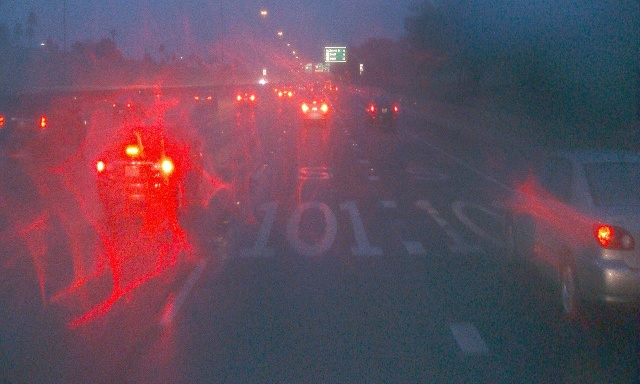}}\\[-2.2ex]

    \subfloat {\raisebox{1.2em}{\rotatebox[origin=t]{90}{\notsotiny ILD}}}\hspace{-0.01cm}\,
    \subfloat {\includegraphics[width=2.42cm,height=1.1cm]{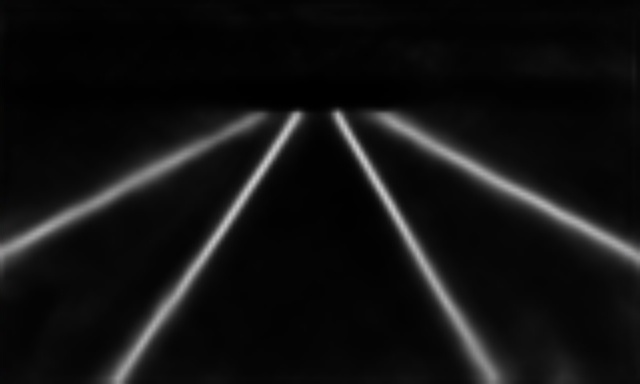}}\,\!\!
    \subfloat {\includegraphics[width=2.42cm,height=1.1cm]{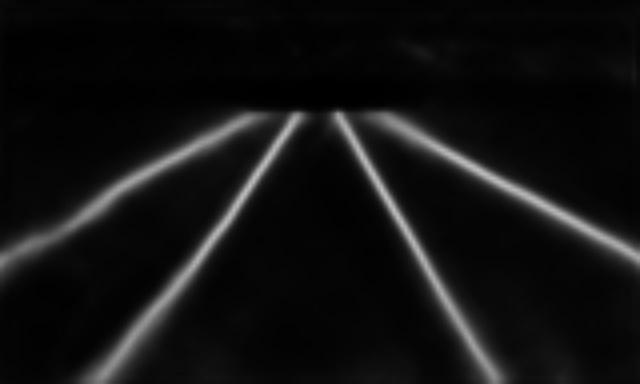}}\,\!\!
    \subfloat {\includegraphics[width=2.42cm,height=1.1cm]{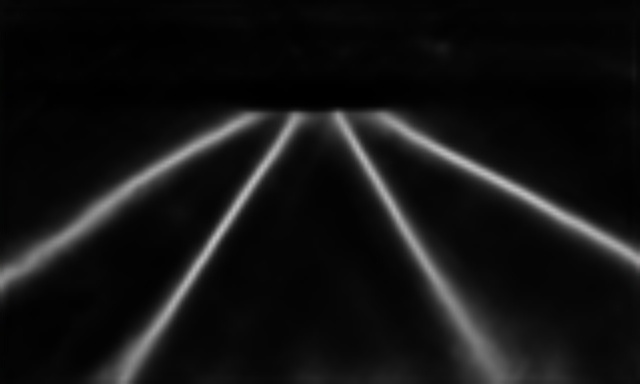}}\,\!\!
    \subfloat {\includegraphics[width=2.42cm,height=1.1cm]{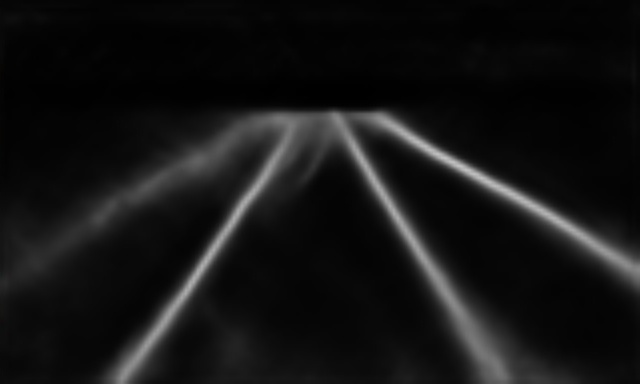}}\,\!\!
    \subfloat {\includegraphics[width=2.42cm,height=1.1cm]{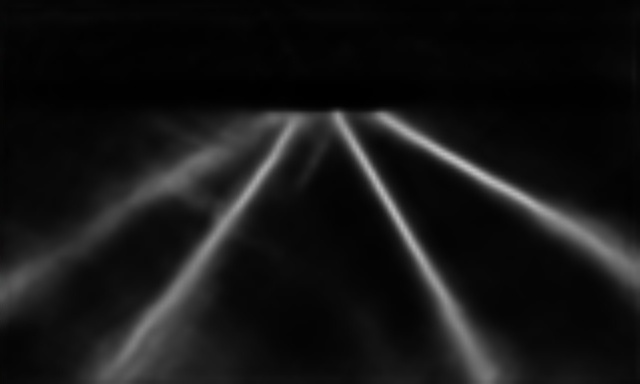}}\,\!\!
    \subfloat {\includegraphics[width=2.42cm,height=1.1cm]{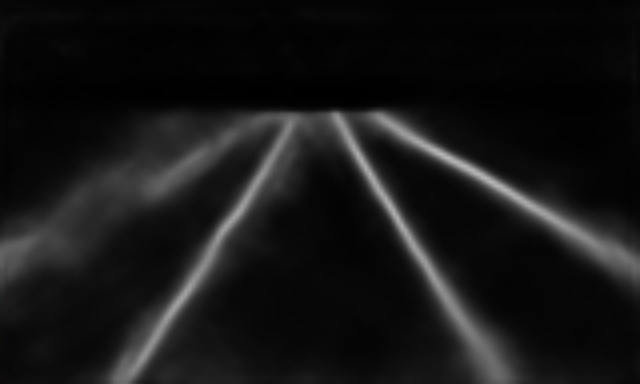}}\,\!\!
    \subfloat {\includegraphics[width=2.42cm,height=1.1cm]{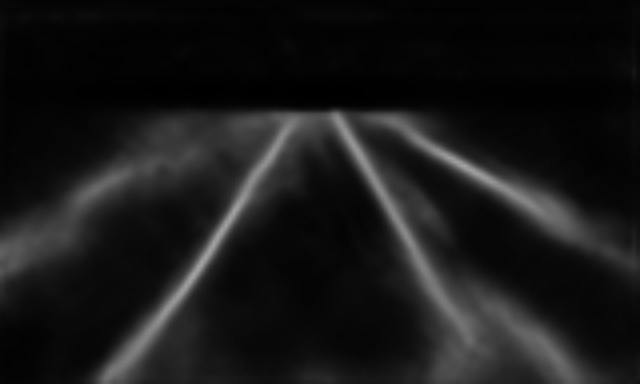}}\\[-2.2ex]

    \setcounter{subfigure}{-1}

    \subfloat {\raisebox{1.2em}{\rotatebox[origin=t]{90}{\notsotiny RVLD}}}\hspace{-0.01cm}\,
    \subfloat {\includegraphics[width=2.42cm,height=1.1cm]{figure/result/stable_results/selected/ILD/v1/7}}\,\!\!
    \subfloat {\includegraphics[width=2.42cm,height=1.1cm]{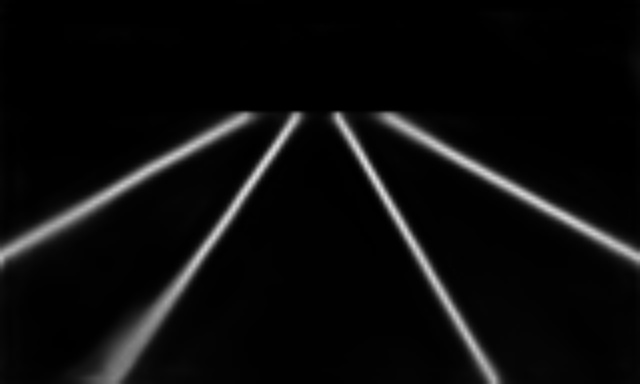}}\,\!\!
    \subfloat {\includegraphics[width=2.42cm,height=1.1cm]{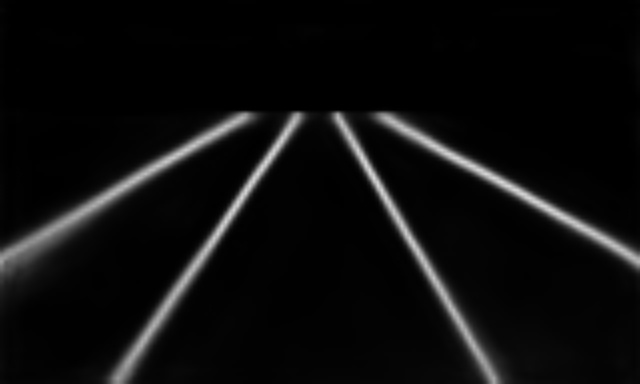}}\,\!\!
    \subfloat {\includegraphics[width=2.42cm,height=1.1cm]{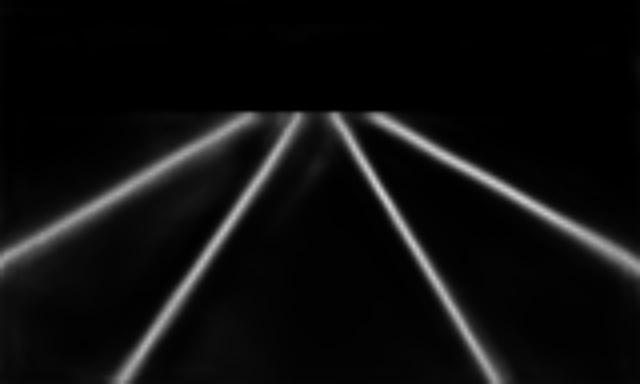}}\,\!\!
    \subfloat {\includegraphics[width=2.42cm,height=1.1cm]{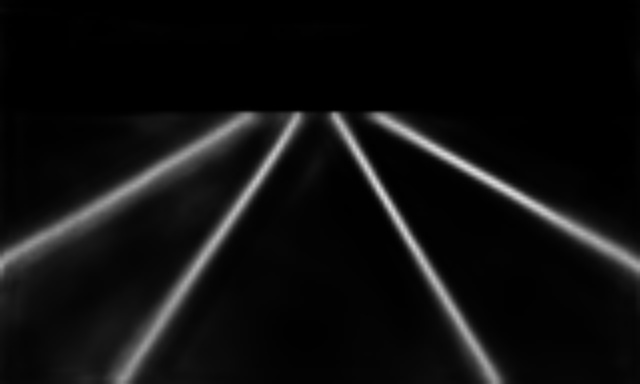}}\,\!\!
    \subfloat {\includegraphics[width=2.42cm,height=1.1cm]{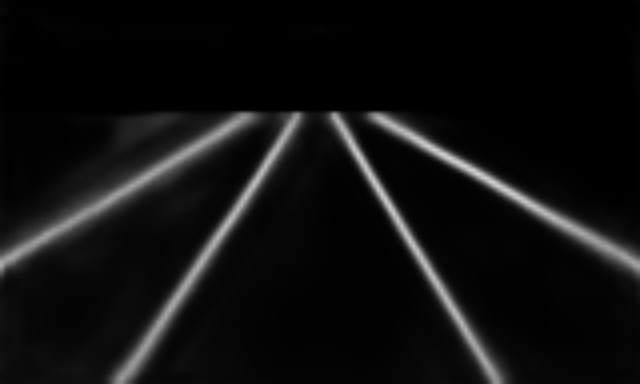}}\,\!\!
    \subfloat {\includegraphics[width=2.42cm,height=1.1cm]{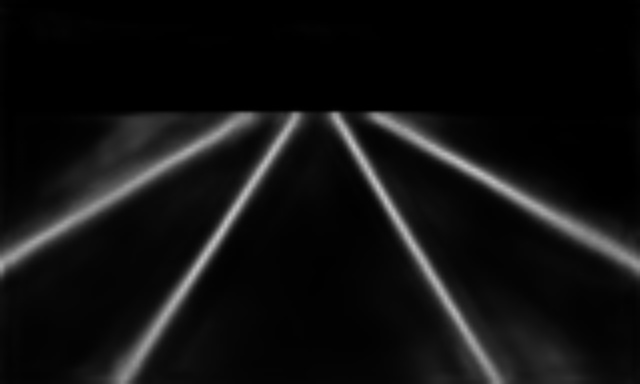}}\\[-1.8ex]

    \subfloat {\raisebox{1.2em}{\rotatebox[origin=t]{90}{\notsotiny Image}}}\hspace{-0.01cm}\,\!
    \subfloat {\includegraphics[width=2.42cm,height=1.1cm]{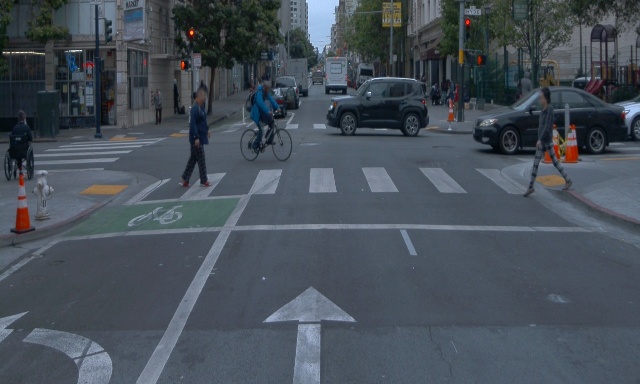}}\,\!\!
    \subfloat {\includegraphics[width=2.42cm,height=1.1cm]{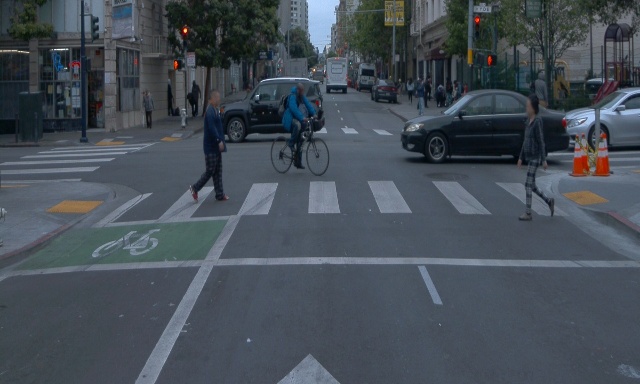}}\,\!\!
    \subfloat {\includegraphics[width=2.42cm,height=1.1cm]{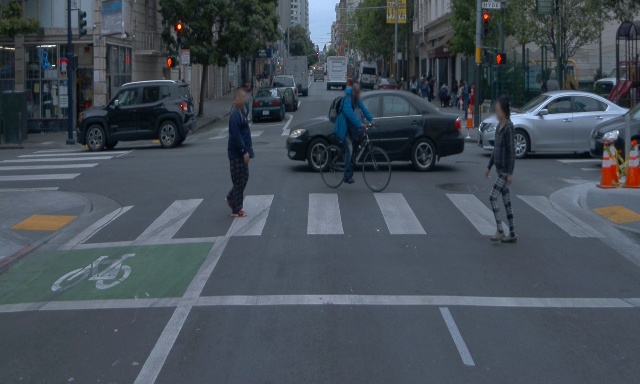}}\,\!\!
    \subfloat {\includegraphics[width=2.42cm,height=1.1cm]{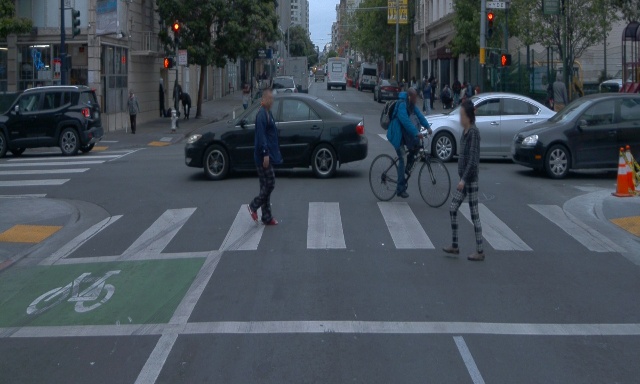}}\,\!\!
    \subfloat {\includegraphics[width=2.42cm,height=1.1cm]{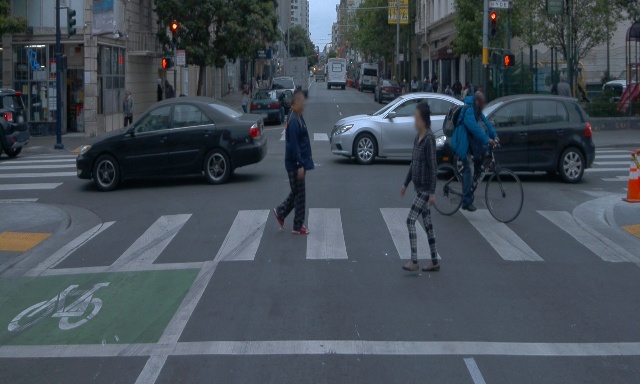}}\,\!\!
    \subfloat {\includegraphics[width=2.42cm,height=1.1cm]{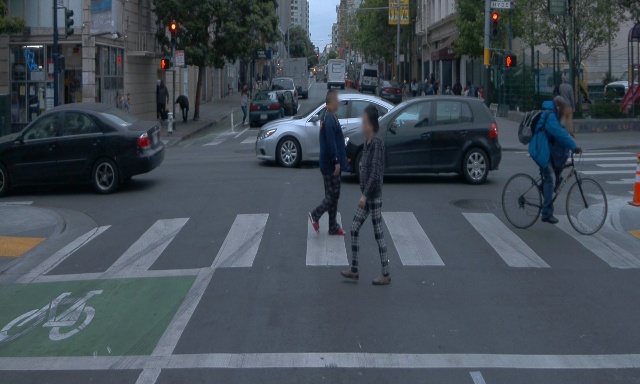}}\,\!\!
    \subfloat {\includegraphics[width=2.42cm,height=1.1cm]{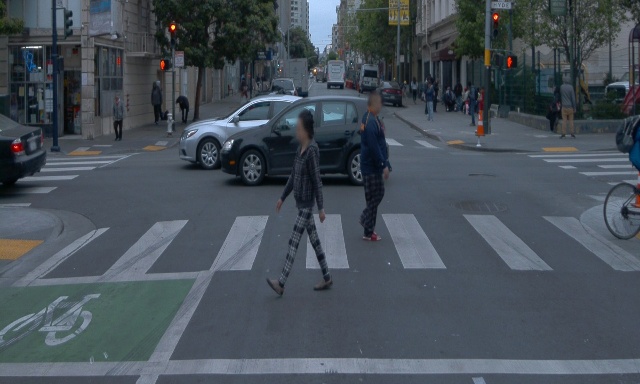}}\\[-2.2ex]

    \setcounter{subfigure}{-1}
    \subfloat {\raisebox{1.2em}{\rotatebox[origin=t]{90}{\notsotiny ILD}}}\hspace{-0.01cm}\,
    \subfloat {\includegraphics[width=2.42cm,height=1.1cm]{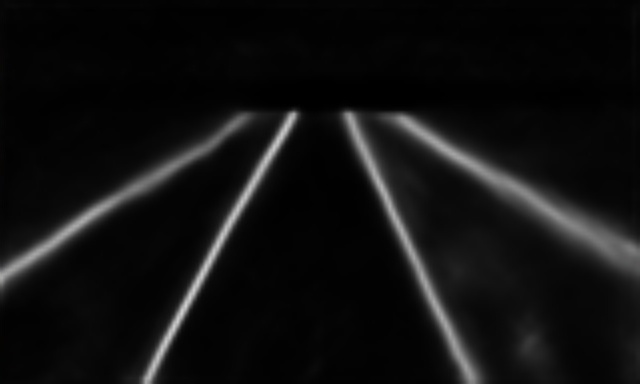}}\,\!\!
    \subfloat {\includegraphics[width=2.42cm,height=1.1cm]{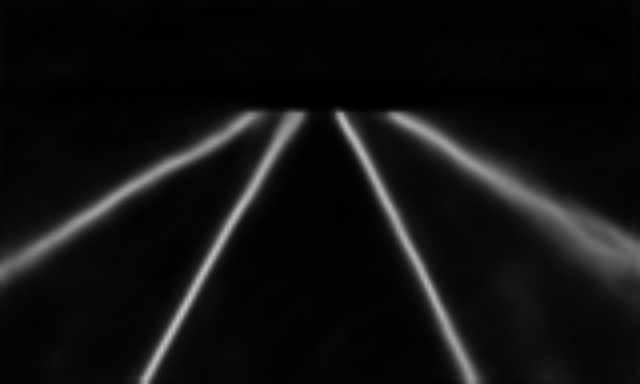}}\,\!\!
    \subfloat {\includegraphics[width=2.42cm,height=1.1cm]{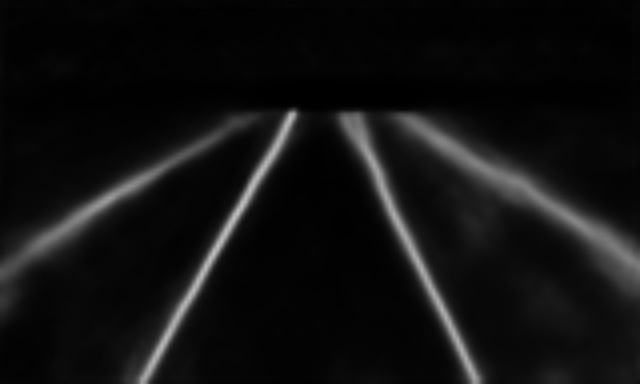}}\,\!\!
    \subfloat {\includegraphics[width=2.42cm,height=1.1cm]{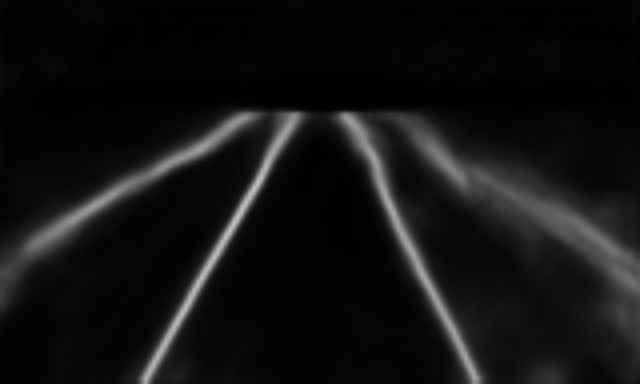}}\,\!\!
    \subfloat {\includegraphics[width=2.42cm,height=1.1cm]{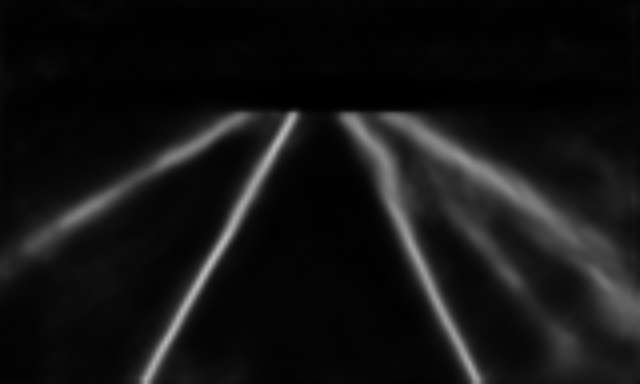}}\,\!\!
    \subfloat {\includegraphics[width=2.42cm,height=1.1cm]{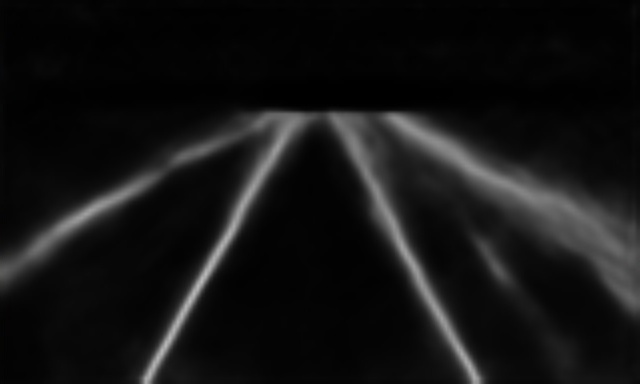}}\,\!\!
    \subfloat {\includegraphics[width=2.42cm,height=1.1cm]{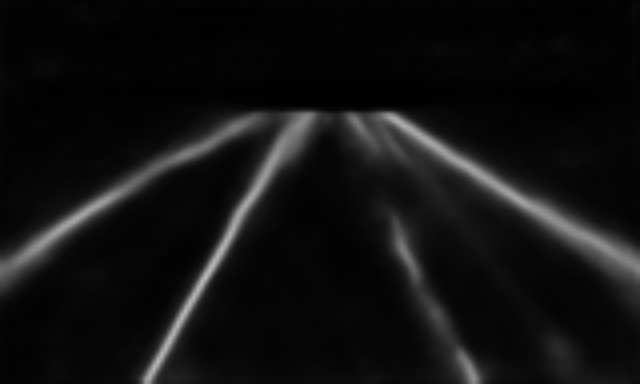}}\\[-2.2ex]

    \setcounter{subfigure}{-1}

    \subfloat {\raisebox{1.2em}{\rotatebox[origin=t]{90}{\notsotiny RVLD}}}\hspace{-0.01cm}\,
    \subfloat {\includegraphics[width=2.42cm,height=1.1cm]{figure/result/stable_results/selected/ILD/v2/0}}\,\!\!
    \subfloat {\includegraphics[width=2.42cm,height=1.1cm]{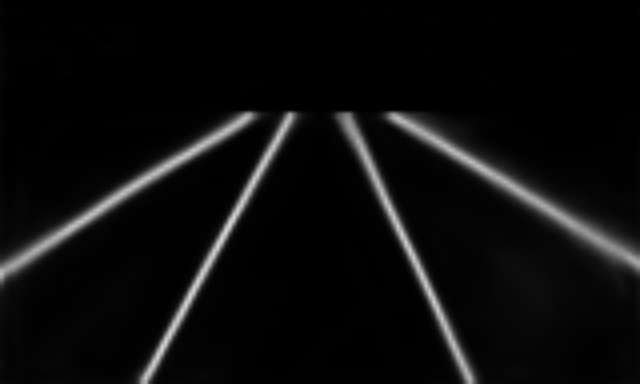}}\,\!\!
    \subfloat {\includegraphics[width=2.42cm,height=1.1cm]{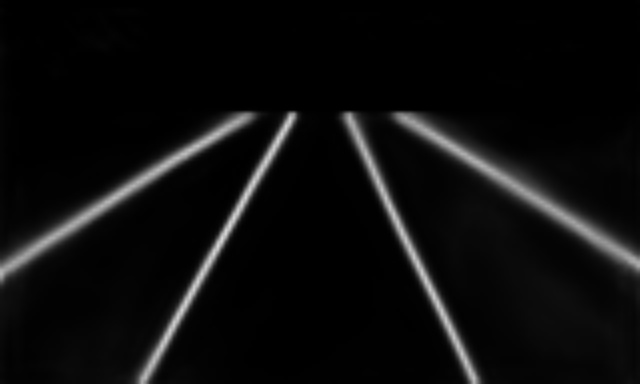}}\,\!\!
    \subfloat {\includegraphics[width=2.42cm,height=1.1cm]{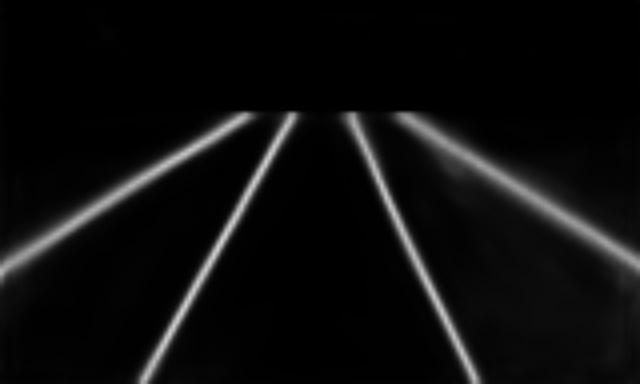}}\,\!\!
    \subfloat {\includegraphics[width=2.42cm,height=1.1cm]{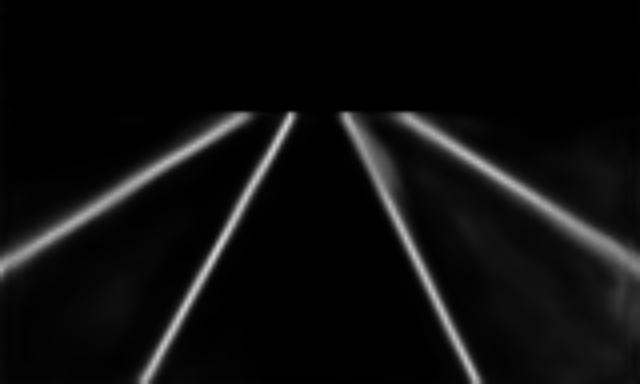}}\,\!\!
    \subfloat {\includegraphics[width=2.42cm,height=1.1cm]{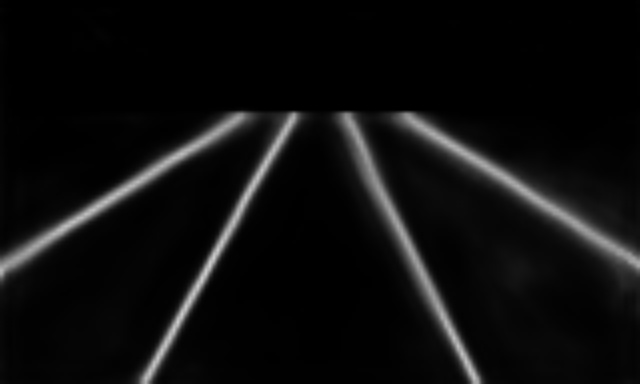}}\,\!\!
    \subfloat {\includegraphics[width=2.42cm,height=1.1cm]{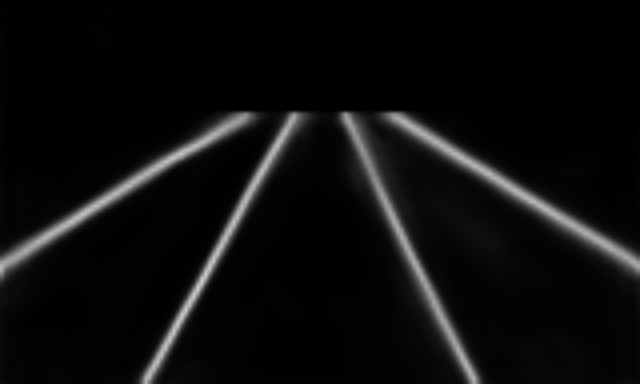}}

    \caption
    {
        Comparison of the lane probability maps generated by ILD and RVLD for consecutive video frames in the OpenLane-V dataset. It is recommended to see the accompanying video clip for more comparisons of lane probability maps.
    }
    \label{fig:ILDvsPLD}
    \end{flushright}
    \vspace*{-0.4cm}
\end{figure*}
\captionsetup[subfigure]{labelformat=parens}

\vspace*{0.1cm}
\noindent\textbf{OpenLane-V:}
Table~\ref{table:openlane} compares RVLD with the state-of-the-art image lane detectors \cite{liu2021condlanenet,qiu2022mfialane,wang2022,zheng2022} and the video lane detectors \cite{zou2019robust,zhang2021lane,zhang2021} on OpenLane-V. Notice that MFIALane \cite{qiu2022mfialane} ranks 2nd on VIL-100 in Table~\ref{table:vil100img}. Also, CondLaneNet \cite{liu2021condlanenet}, GANet \cite{wang2022}, and CLRNet \cite{zheng2022} are recent detectors achieving outstanding performances on image datasets. We train all these detectors on OpenLane-V using publicly available source codes.

In Table~\ref{table:openlane}, we see that RVLD outperforms the existing techniques in every metric, except that it yields a slightly lower mIoU score than CLRNet. GANet and CLRNet yield decent results in terms of image-based mIoU and F1 scores, but they perform poorly in terms of video-based flickering and missing rates. Especially, their ${\rm R}_{\rm F}^{0.5}$ scores are over 0.044, which are at least three times higher than that of RVLD. In other words, they provide temporally less stable lane detection results than RVLD, as also supported by Figure~\ref{fig:graph}. Moreover, RVLD provides better results than the existing video lane detectors, which are ConvLSTM, ConvGRUs, and MMA-Net. Note that RVLD uses a single previous frame only to improve detection results, whereas the existing methods do several past frames. This confirms that the proposed RVLD exploits temporal correlation more effectively. The efficacy of using a single previous frame is demonstrated in the supplement (Section C).

Figure~\ref{fig:openlane_result} shows detection results. The state-of-the-art image-based techniques inaccurately detect implied lanes or simply miss them in challenging scenes. In contrast, RVLD detects the lanes reliably. More detection results are presented in the supplemental document (Section D).

\subsection{Ablation Studies}
We conduct ablation studies to analyze the efficacy of the proposed RVLD and its components. Table~\ref{table:PLD} compares several ablated methods on OpenLane-V. Method \RomNum{1} detects road lanes in each frame using ILD only, without exploiting temporal correlation in a video. In Methods \RomNum{2}$\sim$\RomNum{5}, both ILD and PLD are employed: Method \RomNum{2} does not warp the previous output and uses it directly to refine the feature map of a current frame. Method \RomNum{3} does not use the guidance feature map $G^{t-1}$ in \eqref{eq:agg_refine}.
Method \RomNum{4} does not reuse the refined feature in the future frame. Specifically, in Figure~\ref{fig:Overview}(b), it passes the ILD feature $\tilde{X}^t$, instead of the refined feature $X^t$, to the future frame.

\vspace*{0.1cm}
\noindent\textbf{Efficacy of PLD:} Method~\RomNum{1} underperforms badly. As compared with the proposed RVLD (Method \RomNum{5}), its $\rm F1$ score is reduced from 0.822 to 0.784, and its flickering rate ${\rm R}_{\rm F}^{0.5}$ is about 4 times higher. Using ILD only, it fails to detect unobvious lanes reliably and suffers from temporal instability.

Figure~\ref{fig:ILDvsPLD} compares the lane probability maps of ILD and RVLD on challenging videos in OpenLane-V. As lanes vanish or gradually disappear in videos, ILD yields highly unstable results. On the contrary, RVLD estimates the probability maps clearly and stably, for it refines the feature map of each frame using the past output and propagates the refined map to the future frame.

\vspace*{0.1cm}
\noindent\textbf{Efficacy of motion estimation:}
Compared to RVLD, Method \RomNum{2} yields inferior results, indicating that RVLD can exploit temporal correlation more effectively via the motion-based alignment of past information.

\begin{table}[t]\centering
    \renewcommand{\arraystretch}{0.95}
    \caption
    {
        Ablation studies of the proposed RVLD on OpenLane-V.
    }
    \vspace*{-0.15cm}
    \resizebox{0.9\linewidth}{!}{
    \begin{tabular}[t]{+C{0.2cm}^L{3.2cm}^C{1.2cm}^C{1.2cm}^C{1.2cm}}
    \toprule
    & & ${\rm F1}^{0.5}$ & ${\rm R}_{\rm F}^{0.5}$ & ${\rm R}_{\rm M}^{0.5}$ \\
    \midrule
         \RomNum{1}. & ILD                                  & 0.787 & 0.053 & 0.197 \\
         \RomNum{2}. & ILD + PLD {\scriptsize (w/o warping)}   & 0.816 & 0.017 & 0.191\\
         \RomNum{3}. & ILD + PLD {\scriptsize (w/o guidance)}   & 0.821 & 0.027 & 0.187\\
         \RomNum{4}. & ILD + PLD {\scriptsize (w/o reuse)}   & 0.822 & 0.017 & 0.172\\
         \RomNum{5}. & RVLD                                 & 0.825 & 0.014 & 0.167\\

    \bottomrule
    \end{tabular}}
    \vspace*{-0.25cm}
    \label{table:PLD}
\end{table}

\vspace*{0.1cm}
\noindent\textbf{Efficacy of guidance information:}
Without guidance features in Method~\RomNum{3}, the flickering rate ${\rm R}_{\rm F}^{0.5}$ nearly doubles. Guidance features are essential for stable lane detection, especially when lanes are partially occluded, since they help to maintain lane continuity.

\vspace*{0.1cm}
\noindent\textbf{Efficacy of feature reuse:}
Compared to Method \RomNum{5}, Method \RomNum{4} yields poorer results, indicating that it is more effective to pass the PLD feature $X^t$, rather than the ILD feature $\tilde{X}^t$, to the future frame. In other words, it is better to reuse the refinement feature in the future frame.

\begin{table}[t]\centering
    \renewcommand{\arraystretch}{0.95}
    \caption
    {
        Runtime analysis of the proposed RVLD. MEFR means the motion estimation and feature refinement processes in PLD. The processing times are reported in seconds per frame.
    }
    \vspace*{-0.15cm}
    \resizebox{0.95\linewidth}{!}{
    \begin{tabular}[t]{C{1.5cm}^C{1.5cm}^C{1.5cm}^C{1.5cm}^C{1.5cm}}
    \toprule
    Encoding & MEFR & Decoding & NMS & Total\\
    \midrule
     0.0060s & 0.0034s & 0.0011s & 0.0020s & 0.0125s \\
    \bottomrule
    \end{tabular}}
    \label{table:runtime}
\end{table}

\vspace*{0.1cm}
\noindent\textbf{Runtime:} Table~\ref{table:runtime} lists the runtime for each stage of RVLD. We use a PC with AMD Ryzen 9 5900X CPU and NVIDIA RTX 3090 GPU. The processing speed of RVLD is about 80 frames per second, which is sufficiently fast for applications. ILD, excluding the motion estimation and feature refinement, is faster, but it is less accurate and suffers from temporal incoherence of detected lanes.

\section{Conclusions}
We proposed a novel video lane detector, called RVLD, which extracts informative features for a current frame and passes them recursively to the next frame. First, we designed ILD to localize lanes in a still image. Second, we developed PLD to exploit past information to detect lanes in a current frame more reliably. Experimental results demonstrated that RVLD outperforms existing techniques meaningfully. Moreover, we modified OpenLane to construct OpenLane-V, which is about 10 times larger than VIL-100, and proposed two new video-based metrics, the flickering rate ${\rm R}_{\rm F}$ and the missing rate ${\rm R}_{\rm M}$.

\section*{Acknowledgements}

This work was conducted by Center for Applied Research in Artificial Intelligence (CARAI) grant funded by DAPA and ADD (UD230017TD) and also supported by the NRF grants funded by the Korea government (MSIT) (No.~NRF-2021R1A4A1031864 and No.~NRF-2022R1A2B5B03002310).

\clearpage

{\small
\bibliographystyle{ieeetr}
\bibliography{2023_ICCV_DKJIN}

\begin{thebibliography}{10}

\bibitem{he2004}
Y.~He, H.~Wang, and B.~Zhang, ``Color-based road detection in urban traffic
  scenes,'' {\em IEEE Trans. Intel. Transp. Syst.}, vol.~5, no.~4,
  pp.~309--318, 2004.

\bibitem{aly2008}
M.~Aly, ``Real time detection of lane markers in urban streets,'' in {\em
  Intelligent Vehicles Symposium}, 2008.

\bibitem{hillel2014}
A.~B. Hillel, R.~Lerner, D.~Levi, and G.~Raz, ``Recent progresss in road and
  lane detection{:} {A} survey,'' {\em Mach Vis. Appl.}, vol.~25, no.~3,
  pp.~727--745, 2014.

\bibitem{zhou2010}
S.~Zhou, Y.~Jiang, J.~Xi, J.~Gong, G.~Xiong, and H.~Chen, ``A novel lane
  detection based on geometrical model and {G}abor filter,'' in {\em
  Intelligent Vehicles Symposium}, 2010.

\bibitem{pan2018}
X.~Pan, J.~Shi, P.~Luo, X.~Wang, and X.~Tang, ``Spatial as deep: {S}patial
  {CNN} for traffic scene understanding,'' in {\em Proc. AAAI}, 2018.

\bibitem{zheng2021}
T.~Zheng, H.~Fang, Y.~Zhang, W.~Tang, Z.~Yang, H.~Liu, and D.~Cai, ``{RESA}:
  {R}ecurrent feature-shift aggregator for lane detection,'' in {\em Proc.
  AAAI}, 2021.

\bibitem{qiu2022mfialane}
Z.~Qiu, J.~Zhao, and S.~Sun, ``{MFIAL}ane: {M}ultiscale feature information
  aggregator network for lane detection,'' {\em IEEE Trans. Intel. Transp.
  Syst.}, 2022.

\bibitem{hou2019road}
Y.~Hou, Z.~Ma, C.~Liu, and C.~C. Loy, ``Learning lightweight lane detection
  {CNN}s by self attention distillation,'' in {\em Proc. IEEE ICCV}, 2019.

\bibitem{hou2020inter}
Y.~Hou, Z.~Ma, C.~Liu, T.-W. Hui, and C.~C. Loy, ``Inter-region affinity
  distillation for road marking segmentation,'' in {\em Proc. IEEE CVPR}, 2020.

\bibitem{neven2018}
D.~Neven, B.~De~Brabandere, S.~Georgoulis, M.~Proesmans, and L.~Van~Gool,
  ``Towards end-to-end lane detection: An instance segmentation approach,'' in
  {\em Intelligent Vehicles Symposium}, 2018.

\bibitem{wang2020poly}
B.~Wang, Z.~Wang, and Y.~Zhang, ``Polynomial regression network for
  variable-number lane detection,'' in {\em Proc. ECCV}, 2020.

\bibitem{tabelini2021ICPR}
L.~Tabelini, R.~Berriel, T.~M. Paixao, C.~Badue, A.~F. De~Souza, and
  T.~Oliveira-Santos, ``Poly{L}ane{N}et: {L}ane estimation via deep polynomial
  regression,'' in {\em Proc. IEEE ICPR}, 2021.

\bibitem{liu2021end}
R.~Liu, Z.~Yuan, T.~Liu, and Z.~Xiong, ``End-to-end lane shape prediction with
  transformers,'' in {\em Proc. IEEE WACV}, 2021.

\bibitem{feng2022}
Z.~Feng, S.~Guo, X.~Tan, K.~Xu, M.~Wang, and L.~Ma, ``Rethinking efficient lane
  detection via curve modeling,'' in {\em Proc. IEEE CVPR}, 2022.

\bibitem{qu2021}
Z.~Qu, H.~Jin, Y.~Zhou, Z.~Yang, and W.~Zhang, ``Focus on local: {D}etecting
  lane marker from bottom up via key point,'' in {\em Proc. IEEE CVPR}, 2021.

\bibitem{wang2022}
J.~Wang, Y.~Ma, S.~Huang, T.~Hui, F.~Wang, C.~Qian, and T.~Zhang, ``A
  keypoint-based global association network for lane detection,'' in {\em Proc.
  IEEE CVPR}, 2022.

\bibitem{xu2022}
S.~Xu, X.~Cai, B.~Zhao, L.~Zhang, H.~Xu, Y.~Fu, and X.~Xue, ``{RCL}ane: {R}elay
  chain prediction for lane detection,'' in {\em Proc. ECCV}, 2022.

\bibitem{li2019line}
X.~Li, J.~Li, X.~Hu, and J.~Yang, ``Line-{CNN}: {E}nd-to-end traffic line
  detection with line proposal unit,'' {\em IEEE Trans. Intel. Transp. Syst.},
  vol.~21, no.~1, pp.~248--258, 2019.

\bibitem{tabelini2021CVPR}
L.~Tabelini, R.~Berriel, T.~M. Paixao, C.~Badue, A.~F. De~Souza, and
  T.~Oliveira-Santos, ``Keep your eyes on the lane: Real-time attention-guided
  lane detection,'' in {\em Proc. IEEE CVPR}, 2021.

\bibitem{jin2022}
D.~Jin, W.~Park, S.-G. Jeong, H.~Kwon, and C.-S. Kim, ``Eigenlanes:
  {D}ata-driven lane descriptors for structurally diverse lanes,'' in {\em
  Proc. IEEE CVPR}, 2022.

\bibitem{zheng2022}
T.~Zheng, Y.~Huang, Y.~Liu, W.~Tang, Z.~Yang, D.~Cai, and X.~He, ``{CLRN}et:
  {C}ross layer refinement network for lane detection,'' in {\em Proc. IEEE
  CVPR}, 2022.

\bibitem{zou2019robust}
Q.~Zou, H.~Jiang, Q.~Dai, Y.~Yue, L.~Chen, and Q.~Wang, ``Robust lane detection
  from continuous driving scenes using deep neural networks,'' {\em IEEE Trans.
  Veh. Technol.}, vol.~69, no.~1, pp.~41--54, 2019.

\bibitem{zhang2021lane}
J.~Zhang, T.~Deng, F.~Yan, and W.~Liu, ``Lane detection model based on
  spatio-temporal network with double convolutional gated recurrent units,''
  {\em IEEE Trans. Intel. Transp. Syst.}, vol.~23, no.~7, pp.~6666--6678, 2021.

\bibitem{zhang2021}
Y.~Zhang, L.~Zhu, W.~Feng, H.~Fu, M.~Wang, Q.~Li, C.~Li, and S.~Wang,
  ``{VIL}-100: {A} new dataset and a baseline model for video instance lane
  detection,'' in {\em Proc. IEEE ICCV}, 2021.

\bibitem{tabelini2022}
L.~Tabelini, R.~Berriel, A.~F. De~Souza, C.~Badue, and T.~Oliveira-Santos,
  ``Lane marking detection and classification using spatial-temporal feature
  pooling,'' in {\em International Joint Conference on Neural Networks}, 2022.

\bibitem{wang2022video}
M.~Wang, Y.~Zhang, W.~Feng, L.~Zhu, and S.~Wang, ``Video instance lane
  detection via deep temporal and geometry consistency constraints,'' in {\em
  Proc. ACM Multimedia}, 2022.

\bibitem{chen2022}
L.~Chen, C.~Sima, Y.~Li, Z.~Zheng, J.~Xu, X.~Geng, H.~Li, C.~He, J.~Shi,
  Y.~Qiao, and J.~Yan, ``Pers{F}ormer: 3{D} lane detection via perspective
  transformer and the {OpenLane} benchmark,'' in {\em Proc. ECCV}, 2022.

\bibitem{candes2010power}
E.~J. Cand{\`e}s and T.~Tao, ``The power of convex relaxation: {N}ear-optimal
  matrix completion,'' {\em IEEE Trans. Inf. Theory}, vol.~56, no.~5,
  pp.~2053--2080, 2010.

\bibitem{qin2020}
Z.~Qin, H.~Wang, and X.~Li, ``Ultra fast structure-aware deep lane detection,''
  in {\em Proc. ECCV}, 2020.

\bibitem{liu2021condlanenet}
L.~Liu, X.~Chen, S.~Zhu, and P.~Tan, ``Cond{L}ane{N}et: {A} top-to-down lane
  detection framework based on conditional convolution,'' in {\em Proc. IEEE
  ICCV}, 2021.

\bibitem{chen2019}
Z.~Chen, Q.~Liu, and C.~Lian, ``Point{L}ane{N}et: {E}fficient end-to-end {CNN}s
  for accurate real-time lane detection,'' in {\em Intelligent Vehicles
  Symposium}, 2019.

\bibitem{xu2020curvelane}
H.~Xu, S.~Wang, X.~Cai, W.~Zhang, X.~Liang, and Z.~Li, ``Curve{L}ane-{NAS}:
  {U}nifying lane-sensitive architecture search and adaptive point blending,''
  in {\em Proc. ECCV}, 2020.

\bibitem{oh2019}
S.~W. Oh, J.-Y. Lee, N.~Xu, and S.~J. Kim, ``Video object segmentation using
  space-time memory networks,'' in {\em Proc. IEEE ICCV}, 2019.

\bibitem{vaswani2017}
A.~Vaswani, N.~Shazeer, N.~Parmar, J.~Uszkoreit, L.~Jones, A.~N. Gomez,
  {\L}.~Kaiser, and I.~Polosukhin, ``Attention is all you need,'' in {\em Proc.
  NeurIPS}, 2017.

\bibitem{he2016deep}
K.~He, X.~Zhang, S.~Ren, and J.~Sun, ``Deep residual learning for image
  recognition,'' in {\em Proc. IEEE CVPR}, 2016.

\bibitem{winston2004}
W.~L. Winston and J.~B. Goldberg, {\em Operations Research: Applications and
  Algorithms}, vol.~3.
\newblock Thomson Brooks/Cole Belmont, 2004.

\bibitem{sun2018pwc}
D.~Sun, X.~Yang, M.-Y. Liu, and J.~Kautz, ``{PWC}-{N}et: {CNN}s for optical
  flow using pyramid, warping, and cost volume,'' in {\em Proc. IEEE CVPR},
  2018.

\bibitem{wolberg1990invwarping}
G.~Wolberg, {\em Digital Image Warping}.
\newblock IEEE Computer Society Press, 1990.

\bibitem{lin2017focal}
T.-Y. Lin, P.~Goyal, R.~Girshick, K.~He, and P.~Doll{\'a}r, ``Focal loss for
  dense object detection,'' in {\em Proc. IEEE CVPR}, 2017.

\bibitem{tusimple}
``{TuSimple} benchmark.'' [Online]. Available:
  \url{https://github.com/TuSimple/tusimple-benchmark}.

\bibitem{hastie2015}
T.~Hastie, R.~Mazumder, J.~D. Lee, and R.~Zadeh, ``Matrix completion and
  low-rank {SVD} via fast alternating least squares,'' {\em The Journal of
  Machine Learning Research}, vol.~16, no.~1, pp.~3367--3402, 2015.

\bibitem{jain2013}
P.~Jain, P.~Netrapalli, and S.~Sanghavi, ``Low-rank matrix completion using
  alternating minimization,'' in {\em Proc. the 45th Annual ACM Symp. Theory of
  Computing}, 2013.

\bibitem{liu2021}
R.~Liu, Z.~Yuan, T.~Liu, and Z.~Xiong, ``End-to-end lane shape prediction with
  transformers,'' in {\em Proc. IEEE WACV}, 2021.

\end{thebibliography}
}

\end{document}